\documentclass[sn-basic,iicol]{sn-jnl}

\usepackage{graphicx}%
\usepackage{multirow}%
\usepackage{amsmath,amssymb,amsfonts}%
\usepackage{amsthm}%
\usepackage{mathrsfs}%
\usepackage[title]{appendix}%
\usepackage{xcolor}%
\usepackage{textcomp}%
\usepackage{manyfoot}%
\usepackage{booktabs}%
\usepackage{algorithm}%
\usepackage{algorithmicx}%
\usepackage{algpseudocode}%
\usepackage{listings}%

\theoremstyle{thmstyleone}%
\theoremstyle{thmstyletwo}%

\theoremstyle{thmstylethree}%

\begin{document}

\title[Article Title]{Hyperbolic Face Anti-Spoofing}


\author[1,6]{\fnm{Shuangpeng} \sur{Han}}\email{clydehsp98@gmail.com}
\equalcont{These authors contributed equally to this work.}

\author[1]{\fnm{Rizhao} \sur{Cai}}\email{rzcai@ntu.edu.sg}
\equalcont{These authors contributed equally to this work.}

\author[2]{\fnm{Yawen} \sur{Cui}}\email{yawen.cui@oulu.fi}

\author*[3,4]{\fnm{Zitong} \sur{Yu}}\email{yuzitong@gbu.edu.cn}

\author[5]{\fnm{Yongjian} \sur{Hu}}\email{eeyjhu@scut.edu.cn}

\author[1]{\fnm{Alex} \sur{Kot}}\email{eackot@ntu.edu.sg}

\affil[1]{\orgname{Nanyang Technological University}, \country{Singapore}}

\affil[2]{\orgname{University of Oulu}, \country{Finland}}

\affil[3]{\orgname{Great Bay University}, \country{China}}

\affil[4]{\orgname{Great Bay Institute for Advanced Study}, \country{China}}

\affil[5]{\orgname{South China University of Technology}, \country{China}}

\affil[6]{\orgname{I2R and CFAR, Agency for Science, Technology and Research (A*STAR)}, \country{Singapore}}

\abstract{Learning generalized face anti-spoofing (FAS) models against presentation attacks is essential for the security of face recognition systems. Previous FAS methods usually encourage models to extract discriminative features, of which the distances within the same class (bonafide or attack) are pushed close while those between bonafide and attack are pulled away. However, these methods are designed based on Euclidean distance, which lacks generalization ability for unseen attack detection due to poor hierarchy embedding ability. According to the evidence that different spoofing attacks are intrinsically hierarchical, we propose to learn richer hierarchical and discriminative spoofing cues in hyperbolic space. Specifically, for unimodal FAS learning, the feature embeddings are projected into the Poincar\'e ball, and then the hyperbolic binary logistic regression layer is cascaded for classification. To further improve generalization, we conduct hyperbolic contrastive learning for the bonafide only while relaxing the constraints on diverse spoofing attacks. To alleviate the vanishing gradient problem in hyperbolic space, a new feature clipping method is proposed to enhance the training stability of hyperbolic models. Besides, we further design a multimodal FAS framework with Euclidean multimodal feature decomposition and hyperbolic multimodal feature fusion \& classification. Extensive experiments on three benchmark datasets (i.e., WMCA, PADISI-Face, and SiW-M) with diverse attack types demonstrate that the proposed method can bring significant improvement compared to the Euclidean baselines on unseen attack detection. In addition, the proposed framework is also generalized well on four benchmark datasets (i.e., MSU-MFSD, IDIAP REPLAY-ATTACK, CASIA-FASD, and OULU-NPU) with a limited number of attack types. } 




\maketitle

\section{Introduction}
Face recognition technology has been widely used in identification verification because of its convenience and remarkable accuracy. However, face recognition systems are vulnerable to presentation attacks (PAs), such as print, replay, and 3D-mask attacks. For securing the face recognition system, the role of face anti-spoofing (FAS) is recognized as a critical problem to solve \cite{yu2022deep}.

\begin{figure*}[t]
  \centering
   \includegraphics[width=1.0\linewidth]{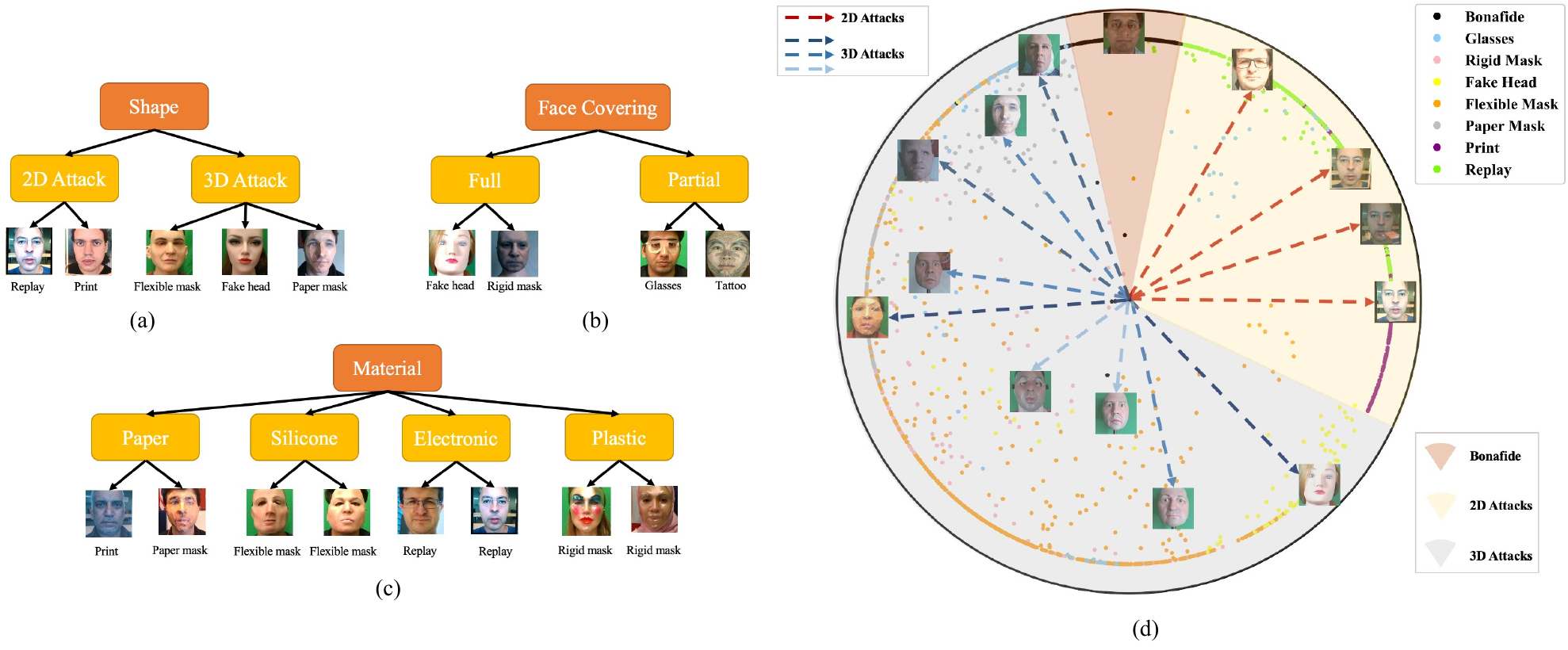}
   \caption{In the FAS data, there are different hierarchical topologies of face spoofing data according to (a) shape; (b) face covering; and (c) material. Hyperbolic space has powerful capabilities of embedding data with many hierarchies. An example of visualization of feature embeddings with Poincar\'e ball model on WMCA~\cite{george2019biometric} testing set under the seen protocol is in (d). Visualization is completed by projecting data into 2-dimensional space via the UMAP algorithm~\cite{mcinnes2018umap}. The closer the sample is to the boundary of the Poincar\'e ball, the easier it is to be classified. The topology of attack types in terms of attack shape is represented by shaded regions. In addition, dotted lines are used to reflect the confidence that the sample is classified. The darker the color of the dotted line, the easier it is for the sample to be classified on the Poincar\'e ball.} 
   \label{fig:Visual_Poincare}
\end{figure*}

In the early stage of the FAS research, traditional methods are mainly using image descriptors to extract hand-craft features~\cite{boulkenafet2015face,Komulainen2014Context}. In the recent era of deep learning, neural networks are used to learn feature representation from data \cite{yu2020multi,yu2020fas,yu2020searching,jourabloo2018face,yang2019face}. While both traditional and deep learning methods can provide saturated performance in intra-domain testing, the cross-domain problem becomes a new challenge urgent to be solved, which means the capturing environments (lighting, cameras, etc.) are different between the training and testing data \cite{FAS-3DCNN-TIFS-2018,shao2019multi}. To tackle the cross-domain problem, FAS methods based on domain generalization have been widely studied \cite{shao2019regularized,metapattern,qin2021meta,wang2022patchnet,SSAN,SSDG-CVPR-2020}, but these works are mainly considering limited 2D attack types (print, replay). As the manufacturing technique develops, more attack types emerge \cite{george2019biometric,DTL}, such as partial mask attacks, silicone masks, glass attacks, etc., bringing new challenges to the FAS problem.

In this work, we focus on improving the generalization capability of FAS models against unseen attack types. To generalize models against unseen attack types, a prevalent solution is to utilize contrastive learning~\cite{liu2022contrastive,sun2022dual,liu2022source}, which pushes close the features' distance of the same class and pulls apart the features' distance of the bonafide and attacks. As such, different attacks, including unseen attacks, are separated from the bonafide in the embedding space. However, learning feature embedding to represent different attacks is non-trivial as there are various types of attacks, such as print attacks, replay attacks, 3D mask attacks, and even partial attacks. Since they have different attributes (as shown in~\autoref{fig:Visual_Poincare}), it is difficult to directly force the different attack types' feature embedding to be close.

 Despite the variety of spoofing attack types, we observe that the different attack types can be categorized hierarchically. As illustrated in~\autoref{fig:Visual_Poincare} (a)(b)(c), PAs might be categorized according to shape (2D or 3D), face covering areas (full or partial), attack materials (e.g., paper, silicone, and plastic), etc. However, a single categorized topology may hardly fit all possible attack types. For example, it is ambiguous to classify half-cropped paper attacks as 2D or 3D attacks. Besides, attack types under different categories may have overlapped attributes. For instance, 2D print paper attack and paper mask attack share the same attributes of paper under the categorization of material, and print paper attack and replay attack share the same attribute of flat shape. Hence, it is hard to describe the intrinsic hierarchical information by human-defined structural topology. 
 
 Motivated by the hierarchical structure of the face spoofing data, we propose to mine the hierarchical information to learn more generalized features in hyperbolic space (as shown in \autoref{fig:Visual_Poincare}). In this paper, we propose to project embeddings into hyperbolic space to mine the hierarchical information to learn generalized features in attack agnostic scenarios. Hyperbolic space can provide powerful representation capability and is suitable for embedding tree-like or hierarchical data~\cite{peng2021hyperbolic,nickel2017poincare,khrulkov2020hyperbolic}. In order to further improve generalization ability, we also propose a contrastive loss in hyperbolic space. In addition, we found that training in hyperbolic space is prone to the problem of vanishing gradient, for which we propose a hyperbolic feature clipping strategy.

After exploring the effect of hyperbolic space embedding on unimodal FAS, we also noted some research work on multimodal FAS~\cite{wang2022conv,liu2021data,yang2020pipenet}. Therefore we try to extend our unimodal hyperbolic FAS framework to multimodal learning. For multimodal FAS, current popular feature fusion strategies include early fusion~\cite{george2020can,wang2022conv} and late fusion ~\cite{shen2019facebagnet,wang2019multi,yang2020pipenet}. 
However, these prevailing methods do not explicitly utilize consistency, complementarity, and compatibility~\cite{guan2021multimodal} among different modal features. These are important properties between modalities, where consistency and complementarity reflect the similarities and differences between different modalities, and compatibility allows different modalities to perform mathematical operations in the same space. Inspired by this, we design a multimodal FAS framework that first decomposes multimodal features into modal-specific and modal-share features which explicitly embody the complementarity and consistency among modalities respectively. Since the distance between modal-share features and unimodal features should be close in the same space, it can be used to supervise the compatibility among modalities. After the feature decomposition, it is necessary to fuse modal-specific and modal-share features and then complete the classification. Due to the powerful representation capability of hyperbolic space embedding, we embed modal-specific and modal-share features into hyperbolic space, and then complete hyperbolic feature fusion and classification. It is worth noting that the feature decomposition process is still completed in Euclidean space, because the feature decomposition operation in hyperbolic space is mathematically complicated, especially hyperbolic concatenation and hyperbolic fully connected layer~\cite{peng2021hyperbolic}, which will not be conducive to the backpropagation of gradients, thus affecting the model learning ability. The main contributions of our work are the following:

 \begin{itemize}
\setlength\itemsep{-0.1em}
     \item We propose to learn discriminative and hierarchical face spoofing cues in hyperbolic space with novel hyperbolic contrastive learning frameworks for unimodal learning and multimodal learning. To our best knowledge, it is the first work to explore the hyperbolic FAS for robust diverse attack detection. 
     \item We introduce a new feature clipping method in hyperbolic space to alleviate the vanishing gradient problem and enhance the training stability of hyperbolic models.
    \item A novel multimodal FAS framework consisting of Euclidean multimodal feature decomposition and hyperbolic multimodal feature fusion \& classification is designed. Consistency, complementarity, and compatibility among different modal features are fully utilized in this framework.
    
   \item  We establish new cross-domain attack-agnostic protocols to evaluate models' generalization capability by conducting cross-dataset testing among WMCA \cite{george2019biometric}, PADISI-Face \cite{rostami2021detection,spinoulas2021multispectral} and SiW-M \cite{DTL} datasets. Proposed models are also tested on the existing cross-dataset protocols of MSU-MFSD~\cite{wen2015face}, IDIAP REPLAY-ATTACK~\cite{chingovska2012effectiveness}, CASIA-FASD~\cite{zhang2012face}, and OULU-NPU~\cite{boulkenafet2017oulu} datasets.

    \item  To demonstrate the powerful generalization ability of hyperbolic multimodal feature fusion \& classification, we compare it with Euclidean multimodal feature fusion \& classification on intra- and inter-dataset protocols of WMCA and PADISI-Face datasets,
    
   \item Proposed methods achieve state-of-the-art unimodal and multimodal performance on intra-dataset unseen attack protocols and newly proposed cross-domain attack-agnostic protocols. 
\end{itemize}

\section{Related Work}
\noindent\textbf{Face Anti-Spoofing.}\quad     
Traditional FAS methods usually extract handcrafted features to train classifiers such as LBP~\cite{boulkenafet2015face} and HOG~\cite{Komulainen2014Context}. More deep learning methods~\cite{liu2021adaptive,zhou2022adaptive,du2022energy,george2020effectiveness} are proposed for face anti-spoofing based on neural networks. The FAS task can be formulated as a binary classification problem~\cite{Li2017An}, and thus binary cross entropy loss is commonly used for model supervision. However, due to the domain shift between different datasets and the difference in attack types, only the binary labels limit the generalization capability of models. To tackle these limitations, auxiliary pixel-wise supervisions~\cite{yu2021revisiting,FAS-Auxiliary-CVPR-2018,george2019deep}, such as pixel maps, and rPPG signals, are utilized for improving the performance. In addition, FAS based on disentangled learning~\cite{liu2020disentangling,zhang2020face}, zero-shot~\cite{DTL,qin2020learning}, anomaly detection~\cite{baweja2020anomaly,fatemifar2021client,li2020unseen,nikisins2018effectiveness}, meta-learning~\cite{chen2021generalizable,liu2021adaptive,liu2021dual,shao2020regularized,wang2021self}, and continuous learning~\cite{perez2020learning,rostami2021detection,cai2023rehearsal} are also explored for domain generalization and adaptation. 

For multimodal FAS learning, different neural network structures and loss functions were researched, such as cross modal focal loss~\cite{george2021cross}, moat loss~\cite{wang2022conv}, modality-agnostic vision transformers~\cite{liu2023ma}, the FAS model incorporating local patch convolution with global MLP~\cite{wang2022conv}, two-stage cascade framework with selective low-level and high-level feature fusion~\cite{liu2021data}, multimodal central difference networks~\cite{yu2020multi}, etc. Besides, among advanced techniques, contrastive learning methods have shown effectiveness. In the paradigm of contrastive learning for FAS~\cite{liu2022contrastive}, features of bonafide are pushed by minimizing the distance, while the features of bonafide and features of attacks are pulled apart by maximizing the margin. 

However, existing works usually calculate the distance in Euclidean space, in which the hierarchy information of different attacks could not be efficiently utilized. In this work, we project feature embedding in hyperbolic space and conduct contrastive learning to learn more generalized features by mining the hierarchy information to tackle the challenges of unseen attack detection.

\noindent\textbf{Hyperbolic Geometry.}\quad
Hyperbolic geometry is non-Euclidean geometry with a constant negative Gaussian curvature. It is pointed out in \cite{khrulkov2020hyperbolic} that semantic similarities and hierarchical relationships between images can be well analyzed in hyperbolic geometry. Also, hyperbolic geometry is beneficial in representing tree-like structures and taxonomies~\cite{ganea2018hyperbolic,nickel2017poincare,nickel2018learning}, text~\cite{aly2019every,tifrea2018poincare}, and
graphs~\cite{bachmann2020constant,liu2019hyperbolic}. There are five isometric models of hyperbolic geometry~\cite{cannon1997hyperbolic}: the Hyperboloid model, the Klein model, the Hemisphere model, the Poincaré ball model, and the Poincaré half-space model. This paper uses the Poincaré ball~\cite{nickel2017poincare} to describe the hyperbolic embedding space. 
Hyperbolic deep neural networks~\cite{nickel2017poincare,ganea2018hyperbolic} are powerful geometrical representations with neural structures built in the space of hyperbolic geometry. Nickel and Kiela~\cite{nickel2017poincare} first learn representation in hyperbolic space. Then, hyperbolic geometry is introduced into deep learning in hyperbolic neural networks~\cite{ganea2018hyperbolic}. Recently, works of hyperbolic image embeddings \cite{cui2022rethinking,khrulkov2020hyperbolic,liu2020hyperbolic} tend to append shallow hyperbolic layers~\cite{ganea2018hyperbolic} after Euclidean deep convolutional neural networks. In this paper, we apply the hyperbolic operation~\cite{khrulkov2020hyperbolic} on FAS for hyperbolic embedding projection.

\section{Methodology} \label{sec:method}
In this section, the preliminary knowledge and the technical details of the hyperbolic FAS framework are introduced. First, we provide preliminary knowledge of hyperbolic geometry and define operations in hyperbolic space. Then, we discuss hyperbolic cross-entropy loss, hyperbolic contrastive loss, and hyperbolic feature clipping for model optimization. Finally, we describe the proposed hyperbolic unimodal face anti-spoofing framework and hyperbolic multimodal face anti-spoofing framework in detail.
\subsection{Preliminary: Hyperbolic Geometry and Operations}
\label{sec:hyp_go}
One main challenge for FAS is to detect a wide variety of seen and unseen attacks. We notice that there are many hierarchies among PAs, which can be embedded with the help of the powerful data hierarchy mining capabilities of hyperbolic space. In hyperbolic geometry, several isometric models are available. Our research is based on the commonly used Poincar\'e ball model~\cite{nickel2017poincare}. Formally, an $n$ dimensional Poincar\'e ball model $(\mathbb{H}^n_c, g^{\mathbb{H}})$ is defined with the manifold $\mathbb{H}^n_c$ as:

\begin{equation} 
	\mathbb{H}^n_c = \{\mathbf{x}\in\mathbb{R}^n:c\left\|\mathbf{x}\right\|^2<1,c\ge 0\},
\end{equation}
and the Riemannian metric $g^\mathbb{H} = {\lambda^c_{\mathbf{x}}}^2g^\mathbb{E}$ in associated tangent space $T_\mathbf{x} $, where $c$ is the curvature of the Poincar\'e ball, $\lambda^c_{\mathbf{x}} = \frac{2}{{1-c\left\|\mathbf{x}\right\|}^2}$ is the conformal factor and $g^\mathbb{E} = \mathbf{I}_n$ is the Euclidean metric tensor \cite{ermolov2022hyperbolic}.  The conformal factor makes the volume of the object in hyperbolic space grow exponentially with the increase of the radius, while the volume of the object in Euclidean space grows only polynomially with the increase of the radius. As the data hierarchy deepens, this property enables hyperbolic space to embed data better than Euclidean space. 

Since hyperbolic space does not meet the conditions of the vector space, it is necessary to redefine the arithmetic operation on hyperbolic space, such as addition, multiplication, etc. Fortunately, M\"obius gyrovector spaces provide us with a way to extend standard operations to hyperbolic space. For the hyperbolic Neural Network, the following hyperbolic operations \cite{khrulkov2020hyperbolic} are crucial. 

\noindent\textbf{M\"obius Addition:} For two points $\mathbf{x}$ and $\mathbf{y}\in\mathbb{H}^n_c$, the M\"obius addition is formulated as:

\begin{equation} 
	\mathbf{x} \oplus_c \mathbf{y}:=\frac{\left(1+2 c\langle\mathbf{x}, \mathbf{y}\rangle+c\|\mathbf{y}\|^2\right) \mathbf{x}+\left(1-c\|\mathbf{x}\|^2\right) \mathbf{y}}{1+2 c\langle\mathbf{x}, \mathbf{y}\rangle+c^2\|\mathbf{x}\|^2\|\mathbf{y}\|^2}.
\end{equation}
\textbf{Hyperbolic Distance:} In hyperbolic geometry, the distance between two points $\mathbf{x}$ and $\mathbf{y}\in\mathbb{H}^n_c$ is formulated as:

\begin{equation} 
D_{h y p}(\mathbf{x}, \mathbf{y})=\frac{2}{\sqrt{c}} \operatorname{arctanh}\left(\sqrt{c}\left\|-\mathbf{x} \oplus_c \mathbf{y}\right\|\right).
\end{equation}
\textbf{Exponential Mapping:} It is used to bijectively map the points on Euclidean space to hyperbolic space, and the mathematical expression $\exp _{\mathbf{x}}^c: \mathbb{R}^n \rightarrow \mathbb{H}_c^n$ is defined in the following manner:
\begin{equation} 
\exp _{\mathbf{x}}^c(\mathbf{v}):=\mathbf{x} \oplus_c\left(\tanh \left(\sqrt{c} \frac{\lambda^c_{\mathbf{x}}\|\mathbf{v}\|}{2}\right) \frac{\mathbf{v}}{\sqrt{c}\|\mathbf{v}\|}\right).
\end{equation}
\noindent\textbf{Hyperbolic Binary Logistic Regression (Hyp-BLR):} For the FAS task, given 2 classes ($k=0$ for attack, $k = 1$ for bonafide) and $ \mathbf{p}_k \in \mathbb{H}_c^n, \mathbf{a}_k \in T_{\mathbf{p}_k} \mathbb{H}_c^n \backslash\{\mathbf{0}\}$, the formal definition of binary logistic regression in the Poincar\'e ball is:

\begin{equation} 
\resizebox{7.0cm}{!}{
$p(y=k \mid \mathbf{x}) \propto \exp \left(\frac{\lambda_{\mathbf{p}_k}^c\left\|\mathbf{a}_k\right\|}{\sqrt{c}} \operatorname{arcsinh}\left(\frac{2 \sqrt{c}\left\langle-\mathbf{p}_k \oplus_c \mathbf{x}, \mathbf{a}_k\right\rangle}{\left(1-c\left\|-\mathbf{p}_k \oplus_c \mathbf{x}\right\|^2\right)\left\|\mathbf{a}_k\right\|}\right)\right)$}.
 \label{Eq.MLR}
\end{equation}

\subsection{Hyperbolic Binary Cross Entropy Loss}
The hyperbolic binary cross entropy loss is based on the logit $g_k$ calculated from Eq.(\ref{Eq.MLR}). The predicted probability of the sample being class k (k=0 for attack class and k = 1 for bonafide class) can be expressed as 

\begin{equation} 
p_k = \frac{g_k}{g_0+g_1},\quad k = 0,1.
 \label{Eq.logit}
\end{equation}
Then the hyperbolic binary cross entropy loss is defined as:
\begin{equation} 
\mathcal{L}^{Hyp-BCE}=-(q \log (p_1)+(1-q) \log (p_0)),
\end{equation}
where $q$ is the label (0 for attack and 1 for bonafide). 

\subsection{Hyperbolic Contrastive Loss}
\label{sec:hyp_contra}
To further enhance the models' discrimination capacity, We design a hyperbolic contrastive loss for supervision. Supposing that there are N bonafide samples: $ \mathbf{X} = \{\mathbf{x_1},\mathbf{x_2},...\mathbf{x_N}\}$,  and M attack samples: $\mathbf{Y}=\{\mathbf{y_1},\mathbf{y_2},...\mathbf{y_M}\}$ in the dataset. The idea of hyperbolic contrastive loss is to bring the samples in each class closer in hyperbolic embedding space, that is, to reduce the hyperbolic distance between bonafide samples and the hyperbolic distance between attack samples at the same time. The hyperbolic contrastive loss can be divided into two parts for FAS. The first part is the hyperbolic contrastive loss between bonafide samples. For a bonafide pair $\mathbf{x_i},\mathbf{x_j}\in \mathbf{X}$, the loss function can be formulated as:
\begin{equation} 
\resizebox{7.0cm}{!}{
		$l_{i, j}^{BF}=-\log \frac{\exp \left(-D_{hyp}\left(\mathbf{x}_i, \mathbf{x}_j\right) / \tau\right)}{\exp \left(-D_{hyp}\left(\mathbf{x}_i, \mathbf{x}_j\right) / \tau\right)+\sum_{t=1}^M \exp \left(-D_{hyp}\left(\mathbf{x}_i, \mathbf{y}_t\right) / \tau\right)}$}
	,
 \label{Eq:loss_contra_pair}
\end{equation}
where the superscript BF stands for bonafide samples, and $\tau$ is the temperature coefficient. The second part is the hyperbolic contrastive loss between attack samples. For an attack pair $\mathbf{y_i},\mathbf{y_j}\in \mathbf{Y} $ , the loss function is:
\begin{equation} 
\resizebox{7.0cm}{!}{$
		l_{i, j}^{ATT}=-\log \frac{\exp \left(-D_{hyp}\left(\mathbf{y}_i, \mathbf{y}_j\right) / \tau\right)}{\exp \left(-D_{hyp}\left(\mathbf{y}_i, \mathbf{y}_j\right) / \tau\right)+\sum_{t=1}^N \exp \left(-D_{hyp}\left(\mathbf{y}_i, \mathbf{x}_t\right) / \tau\right)}$},
\end{equation}
where the superscript ATT stands for attack samples. Hence the total hyperbolic contrastive loss $\mathcal{L}^{BA} $is defined as

\begin{equation} 
	\mathcal{L}^{BA} = \frac{1}{N}\sum_{i=1,i\neq j}^{N} \sum_{j=1}^{N}l_{i, j}^{BF}
	+\frac{1}{M}\sum_{u=1,u\neq v}^{M} \sum_{v=1}^{M}l_{u, v}^{ATT}
	\label{HCL_tot},
\end{equation}
\noindent where the superscript BA stands for bonafide samples and attack samples. However, we argue that it is not very reasonable to shorten the distance between attack samples for binary classification for FAS as different attacks may vary greatly (e.g., print and fake head). Forcing the models to learn some unreasonable and unnecessary features will degrade the models' generalization capabilities. Hence, we design the hyperbolic contrastive loss for bonafide samples only, which can be formulated as:

\begin{equation} 
	\mathcal{L}^{BF} = \frac{1}{N}\sum_{i=1,i\neq j}^{N} \sum_{j=1}^{N}l_{i, j}^{BF}.
 \label{Eq:L_contra_BF}
\end{equation}

\subsection{Hyperbolic Feature Clipping}
\label{sec:hyp_clip}
To ensure the numerical stability of the hyperbolic neural network during training, the maximum norm of points on the Poincar\'e ball is often set to $\frac{1}{\sqrt{c}}\left(1-10^{-3}\right)$. Hyperbolic neural networks often have vanishing gradient problems \cite{guo2021free} during training because when the embeddings are closer to the boundary of the Poincar\'e ball, the gradient is closer to zero.
One of the characteristics of hyperbolic-space-based model is to try to push objects that can be easy to classify to the boundary of the Poincar\'e ball, while objects that are difficult to classify are placed close to the center of the Poincar\'e ball. Therefore, when the model training tries to achieve good performance, it is easy to appear the problem of vanishing gradient. Different from \cite{guo2021free}, we propose to solve the vanishing gradient problem by directly clipping the maximum norm of points on the Poincar\'e ball. This is a straightforward but novel approach for feature clipping, and its mathematical expression is as follows: 
\begin{equation} 
	\left\|\mathbf{x}^H\right\| \leq \frac{1}{\sqrt{c}}\left(1-\alpha\right),
\end{equation}
where $\left\|\mathbf{x}^H\right\|$ represents the norm of the point $\mathbf{x}$ on the Poincar\'e ball $(\mathbb{H}^n_c, g^{\mathbb{H}})$ and hyperparameter $\alpha$ indicates the proportion of radius that is ineffective for the Poincar\'e ball. For example, when $\alpha=0.1$, we only take 90\% of the original Poincar\'e ball radius as the effective radius. In this case, embeddings on the Poincar\'e ball can not be pushed close to the original boundary, so as to alleviate the vanishing gradient problem.

\begin{figure*}[t]
  \centering
   \includegraphics[width=1.0\linewidth]{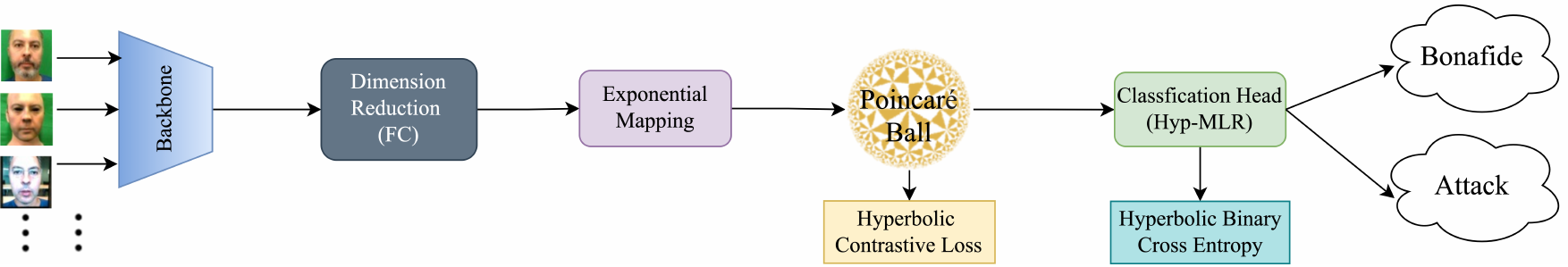}
   \caption{Hyperbolic unimodal FAS framework: Firstly, features extracted by the backbone go through a fully connected (FC) layer for dimension reduction. Then, exponential mapping is utilized to map embeddings from Euclidean space to hyperbolic space, followed by a hyperbolic binary logistic regression layer (Hyp-BLR) for classification. Furthermore, hyperbolic contrastive loss and hyperbolic binary cross-entropy loss are the loss functions during training. }
   \label{fig:SM_frame}
\end{figure*}

\subsection{Hyperbolic Unimodal Face Anti-Spoofing} 
\label{sec:unimodal framework}
The framework of hyperbolic unimodal FAS is shown in Figure \ref{fig:SM_frame}. We first pass feature vectors extracted from the backbone through a fully connected layer to downscale the embedding dimension to reduce the computational cost. Then we project the Euclidean embeddings into hyperbolic embeddings through exponential mapping. Finally, we use a hyperbolic binary logistic regression layer to complete the classification on the Poincar\'e ball. 

The loss function consists of hyperbolic binary cross entropy loss and hyperbolic contrastive loss, which are formulated as:
\begin{equation} 
  \mathcal{L}_{total}^{SM} = \lambda_1\mathcal{L}^{Hyp-BCE} + \lambda_2\mathcal{L}^{BF}.
 \label{Eq:loss_func}
\end{equation}
The ablation study related to the unimodal hyperbolic FAS will be shown in Section~\ref{sec:Ablation}.

\begin{figure*}[t]
  \centering
   \includegraphics[width=1.0\linewidth]{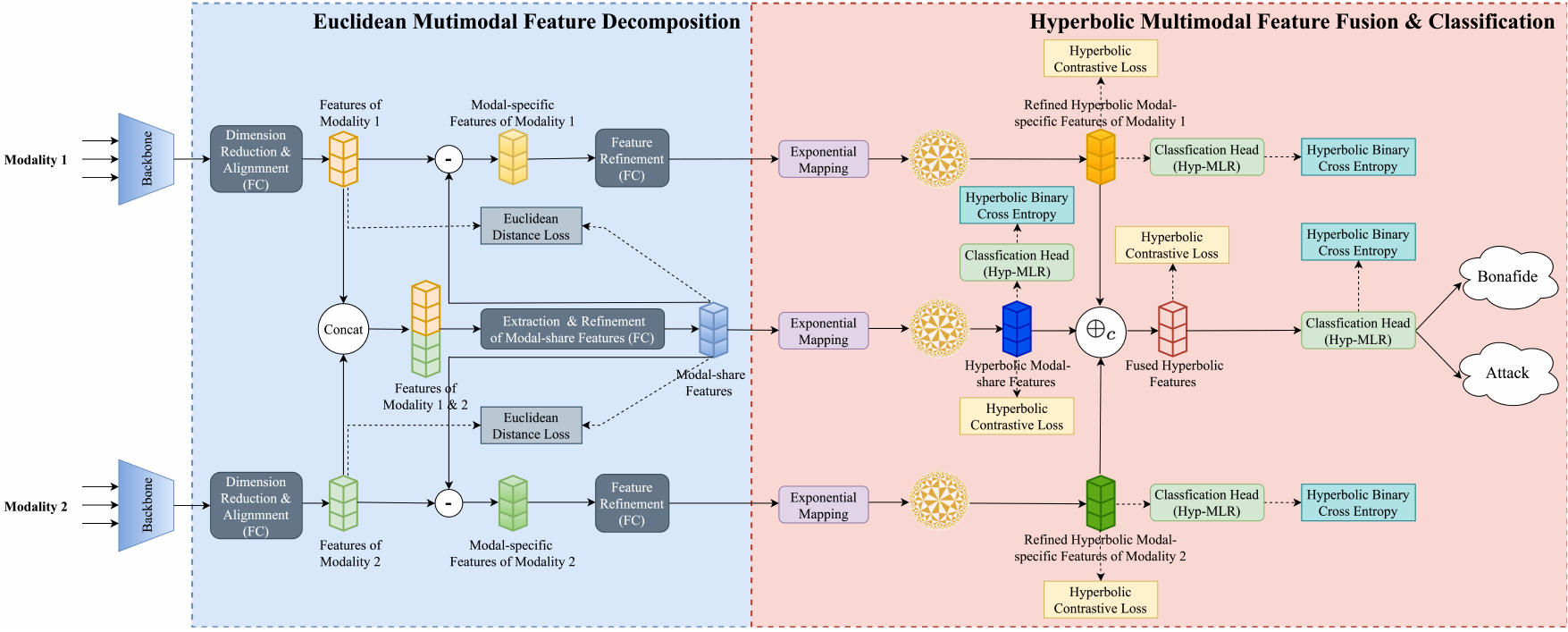}
   \caption{Hyperbolic multimodal FAS framework: Unimodal features extracted by backbones are firstly decomposed in Euclidean space (i.e., Euclidean Multimodal Feature Decomposition), then decomposed features are fused and classified in hyperbolic space (i.e., Hyperbolic Multimodal Feature Fusion \& Classification).  Euclidean distance loss, hyperbolic contrastive loss, and hyperbolic binary cross-entropy loss are utilized as loss functions. FC in this figure refers to the fully connected layer.}
   \label{fig:MM_frame}
\end{figure*}

\subsection{Hyperbolic Multimodal Face Anti-Spoofing} 
\label{sec:multimodal framework}
The unimodal hyperbolic FAS framework can be easily extended to multimodal frameworks such as adding or concatenating unimodal features for classification directly. However, consistency, complementarity and compatibility~\cite{guan2021multimodal} among different modal features are not fully explicitly utilized in this case. Motivated by this, we design a framework that decomposes multimodal features before feature fusion and classification. The framework of hyperbolic multimodal FAS is shown in Figure \ref{fig:MM_frame}. Here we define four terms in our multimodal FAS framework:
\begin{itemize}
    \item \textbf{Euclidean Multimodal Feature Decomposition}: Multimodal features are decomposed in Euclidean space;
    \item \textbf{Hyperbolic Multimodal Feature Fusion \& Classification}: Multimodal features are fused and  then classified in hyperbolic space;
    \item \textbf{Modal-share features}: Information common to all modalities used for learning;
    \item \textbf{Modal-specific features}: Information unique to one modality itself.
\end{itemize}

Formally, we denote two unimodal features extracted by backbones and followed by dimension reduction \& alignment as $\mathcal{F}_1$ and $\mathcal{F}_2$. Then, modal-share features $\mathcal{F}_{sha}$ between $\mathcal{F}_1$ and $\mathcal{F}_2$ can be extracted according to the formula below:
\begin{equation} 
	\mathcal{F}_{sha} = S_{sha}(Concat(\mathcal{F}_1,\mathcal{F}_2),\theta_{sha}),
\end{equation}
where $Concat(\cdot,\cdot)$ stands for concatenation and $S_{sha}(\cdot,\theta_{sha})$ represents modal-share feature extraction and refinement with learnable parameters $\theta_{sha}$. The refinement aims to remove unnecessary and misleading information in modal-share features. In order to supervise that the $\mathcal{F}_{sha}$ can extract the consistency between $\mathcal{F}_1$ and $\mathcal{F}_2$, the Euclidean distance can be utilized as the loss functions:
\begin{equation} 
	\mathcal{L}^{Dis} = \frac{1}{B}\frac{1}{dim}(\left\|\mathcal{F}_{sha}-\mathcal{F}_1\right\|^2+\left\|\mathcal{F}_{sha}-\mathcal{F}_2\right\|^2),
 \label{Eq:L_Dis}
\end{equation}
where $dim$ is the dimension of feature vectors and $B$ is the batch size. The loss $\mathcal{L}^{Dis}$ will make modal-share features and unimodal features close in one space, which guarantees the extraction of consistency among different modal features and the compatibility among $\mathcal{F}_{sha}$, $\mathcal{F}_1$ and $\mathcal{F}_2$. 

The unimodal feature $\mathcal{F}_1$, $\mathcal{F}_2$ can both be decomposed into vector sums of modal-share features and their own modal-specific features:
\begin{equation} 
	\mathcal{F}_{1} = \mathcal{F}_{sha}+\mathcal{F}_{spe1},
\end{equation}
\begin{equation} 
	\mathcal{F}_{2} = \mathcal{F}_{sha}+\mathcal{F}_{spe2}.
\end{equation}
Hence, the modal-specific component of each unimodal feature can be obtained by subtracting $\mathcal{F}_{sha}$ from the corresponding unimodal feature, namely $\mathcal{F}_1-\mathcal{F}_{sha}$ and $\mathcal{F}_2-\mathcal{F}_{sha}$. To remove the redundancy and misleading information, refinements denoted as $S_{spe1}(\cdot,\theta_{spe1})$ and $S_{spe2}(\cdot,\theta_{spe2})$ are applied. Therefore, the refined modal-specific component of $\mathcal{F}_1$ and $\mathcal{F}_2$ are:
\begin{equation} 
	\mathcal{F}_{spe1} = S_{spe1}(\mathcal{F}_1-\mathcal{F}_{sha},\theta_{spe1}),
\end{equation}
\begin{equation} 
	\mathcal{F}_{spe2} = S_{spe2}(\mathcal{F}_2-\mathcal{F}_{sha},\theta_{spe2}).
\end{equation}
 It is worth noting that the extraction of modal-specific features utilizes complementarity among different modal features. At this point, the decomposition of multimodal features has been completed. Consistency, complementarity and compatibility among different modal features are fully used in this decomposition process.

Next, in order to take advantage of the powerful capabilities of feature embedding in hyperbolic space, $\mathcal{F}_{sha}$, $\mathcal{F}_{spe1}$ and $\mathcal{F}_{spe2}$ are mapped into hyperbolic space for multimodal feature fusion and classification, denoted as $\mathcal{F}_{sha}^{Hyp}$, $\mathcal{F}_{spe1}^{Hyp}$ and $\mathcal{F}_{spe2}^{Hyp}$ separately. Their effectiveness is supervised by the hyperbolic binary cross entropy loss ($\mathcal{L}^{Hyp-BCE}_{sha}$, $\mathcal{L}^{Hyp-BCE}_{spe1}$ and $\mathcal{L}^{Hyp-BCE}_{spe2}$) and hyperbolic contrastive loss ($\mathcal{L}^{BF}_{sha}$,$\mathcal{L}^{BF}_{spe1}$ and $\mathcal{L}^{BF}_{spe2}$). Therefore, we denote the total hyperbolic binary cross entropy loss $\mathcal{L}_{decom}^{Hyp-BCE}$ and total hyperbolic contrastive loss $\mathcal{L}_{decom}^{BF}$ of all the features obtained in the decomposition process are:
\begin{equation} 
\resizebox{7.0cm}{!}{
    $\mathcal{L}_{decom}^{Hyp-BCE} = \mathcal{L}^{Hyp-BCE}_{sha}+\mathcal{L}^{Hyp-BCE}_{spe1}+\mathcal{L}^{Hyp-BCE}_{spe2}$},
\end{equation}

\begin{equation} 
    \mathcal{L}_{decom}^{BF} = \mathcal{L}^{BF}_{sha}+\mathcal{L}^{BF}_{spe1}+\mathcal{L}^{BF}_{spe2}.
\end{equation}

Then, these hyperbolic features are fused based on M\"obius addition $\oplus_c$:
\begin{equation} 
    \mathcal{F}_{fus}^{Hyp} = \mathcal{F}_{sha}\oplus_c (\mathcal{F}_{spe1} \oplus_c \mathcal{F}_{spe2}).
\end{equation}

 Finally, the fused feature $\mathcal{F}_{fus}^{Hyp}$ is classified through a hyperbolic binary logistic regression layer. Hyperbolic binary cross entropy loss $\mathcal{L}^{Hyp-BCE}_{fus}$, and hyperbolic contrastive loss $\mathcal{L}^{BF}_{fus}$ are the loss functions of $\mathcal{F}_{fus}^{Hyp}$. Hence, the total loss function of the hyperbolic multimodal face anti-spoofing frame is:
\begin{equation} 
\resizebox{7.0cm}{!}{
    $\mathcal{L}_{total}^{MM} = \gamma_1\mathcal{L}^{Dis}+\gamma_2\mathcal{L}_{decom}^{Hyp-BCE}+ \gamma_3\mathcal{L}_{decom}^{BF}+ \gamma_4\mathcal{L}^{BF}_{fus} +\mathcal{L}^{Hyp-BCE}_{fus}$}.
\end{equation}

The ablation study of the hyperbolic multimodal FAS will be shown in Section~\ref{sec:Ablation}.

\section{Experiments}

\label{sec:experiemnts}
\subsection{Datasets}
To evaluate the FAS models against various attack types, three benchmark datasets (WMCA \cite{george2019biometric}, PADISI-Face \cite{rostami2021detection,spinoulas2021multispectral} and SiW-M \cite{DTL}) with diverse attack types are utilized. The WMCA dataset has overall 1679 short video recordings of both bonafide and presentation attacks, including seven attack types: glasses, fake head, print, replay, rigid mask, flexible mask, and paper mask. It has several modalities, including RGB, depth, infrared, and thermal.  The PADISI-Face dataset contains nine types of attacks: print paper, transparent mask, mannequin (fake head), silicone mask, half mask, makeup, tattoo, funny eye, and paper glasses. It contains multiple modalities: RGB, depth, NIR, thermal, NIRL, NIRR, and SWIR.  The SiW-M dataset collects 968 videos of 13 attack types: replay, print, half mask, silicone mask, transparent mask, papercraft, mannequin, obfuscation, impersonation, cosmetic, funny eye, paper glasses and partial paper. In the multimodal testing, only WMCA and PADISI-Face datasets are utilized since images in the SiW-M dataset are only in RGB format. 

In addition, in order to evaluate the performance of hyperbolic FAS models on datasets with a limited number of attack types, another four benchmark datasets (MSU-MFSD~\cite{wen2015face}(M),
IDIAP REPLAY-ATTACK~\cite{chingovska2012effectiveness}(I), CASIA-FASD~\cite{zhang2012face}(C), and OULU-NPU~\cite{boulkenafet2017oulu}(O)) datasets are adopted. The MSU-MFSD database contains 440 video clips, of which 110 are genuine faces and 330 are spoofing attacks. Attack types in this dataset are printed photos and replayed videos. IDIAP REPLAY-ATTACK contains 1300 video clips of print and replay spoofing attacks. CASIA-FASD collects data with different imaging qualities from 50 genuine subjects. It has three types of attacks: warped photo, cut photo, and video playback. OULU-NPU captures genuine and spoofing faces from 55 different identities, including 2 kinds of print attacks and 2 kinds of video-replay attacks. 

\subsection{Evaluation Protocols}\label{sec:Metrics}
In the intra-dataset experiment of WMCA, we use the grandtest (seen protocol) and leave-one-out (LOO) unseen attack protocols of WCMA. Unlike WMCA, there are no comprehensive unseen attack evaluation protocols for the PADISI-Face dataset. Hence, inspired by the WMCA LOO protocols, we establish LOO protocols for the PADISI-Face dataset to evaluate the model performance against unseen attack types. For the grandtest (seen) protocol of the PADISI-Face dataset, part 0 of the 3-fold PADISI-Face protocol is used. For the SiW-M dataset, LOO protocols are utilized for intra-dataset testing. The Average Classification Error Rate (ACER)~\cite{ACER} is used as the metric to evaluate model performance on the intra-dataset testing on WMCA/PADISI-Face datasets. For the SiW-M dataset, since there are only \textit{train} set and \textit{test} set (no \textit{dev} set) in the LOO protocols, the ACER with a fixed threshold of 0.5 and Equal Error Rate (EER) are used as intra-dataset testing metrics.

Moreover, we propose cross-domain attack-agnostic protocols on cross-dataset testings among WMCA, PADISI-Face and SiW-M datasets to evaluate models' generalization capability against domain shifts (e.g., camera sensors and lightings) and agnostic attacks (e.g., paper mask in WMCA and tattoo in PADISI-Face). The first kind of cross-domain attack-agnostic protocol is to randomly use one of WMCA/PADISI-Face/SiW-M as the \textit{train} set to train the model, and then use one of the remaining two data sets as the \textit{test} set to calculate the metric. Hence, there are six testing tasks in total in this protocol: WMCA to PADISI-Face, WMCA to SiW-M, PADISI-Face to WMCA, PADISI-Face to SiW-M, SiW-M to WMCA, and SiW-M to PADISI-Face. The second kind of cross-domain attack-agnostic protocol is to select two of the WMCA/PADISI-Face/SiW-M datasets as the \textit{train} set, and the remaining dataset as the \textit{test} set for testing. Therefore, WMCA \& PADISI to SiW-M, WMCA \& SiW-M to PADISI-Face, and PADISI-Face \& SiW-M to WMCA are the three tasks. For the datasets with limited attack types, MSU-MFS(M), IDIAP REPLAY-ATTACK(I), CASIA-FASD(C), and OULU-NPU(O), we take two existing protocols M\&I to C and M\&I to O \cite{SSDG-CVPR-2020, cai2022learning} to complete the model evaluation, where two datasets are used for training and one dataset is used for testing. We report the Area Under the Curve (AUC) and Half Total Error Rate (HTER)~\cite{chingovska2014biometrics} with a fixed threshold of 0.5 as metrics to evaluate model performance on cross-dataset protocols.
\begin{table*}[t]
  \caption{Unimodal results of seen and unseen protocols on WMCA dataset. The values ACER(\%) reported on testing sets are obtained with thresholds computed for BPCER=1\% on development sets. The best results are bolded.}
\centering
\setlength{\tabcolsep}{11pt}
  \resizebox{16cm}{!}{
        \begin{tabular}{cccccccccc}
    \toprule[2pt]
    {\textbf{Method}} & {\textbf{Seen}} & \multicolumn{7}{c}{\textbf{Unseen}}                  & {\textbf{Average}} \\
\cmidrule{3-9}          &       & Glasses & Rigid Mask & Fake Head & Flexible Mask & Paper Mask & Print & Replay &  \\
    \midrule[2pt]
    ResNet50 w/CCL~\cite{liu2022contrastive}& 30.69 &  16.80 & 17.62 & 24.67 & \textbf{4.76}  & 9.51 & 19.03 & 15.37   & 15.39$\pm$6.51 \\
    Aux.(Depth) w/CCL~\cite{liu2022contrastive} & 30.62 &  10.17 & 27.32 & 40.00 & 7.41  & 11.67 & 16.11 & 12.76  & 17.92$\pm$11.66 \\
    CDCN w/CCL~\cite{liu2022contrastive} &  27.14 & 35.13 & 15.10 & 21.82  & 7.18& 18.91 & 20.53 & \textbf{11.79}   & 18.64$\pm$8.91 \\
    EPCR~\cite{wang2021consistency} &  - & 41.61 & \textbf{2.78} & 4.97  & 9.91 & 2.47 & \textbf{0.28} & 20.89   & 11.85$\pm$14.85 \\
    \midrule[2pt]
    ViT~\cite{dosovitskiy2020image,liao2023domain}  & 6.89  & 34.52 & 4.86  & 10.01 & 22.56 & 6.07  & 2.65  & 23.11 & 14.83$\pm$12.00 \\
    \textbf{ViT-Hyp (ours)} & 6.34  & 34.47 & 5.40  & 4.00  & 18.46 & 5.10  & 1.13  & 17.90 & 12.35$\pm$11.93 \\
    \textbf{ViT-Hyp-HCL(ours)} & 5.62  & 26.64 & 3.88  & 4.35  & 20.68 & 3.47  & 0.91  & 20.73 & 11.52$\pm$10.68 \\
    \midrule[2pt]
    MCDeepPixBiS~\cite{george2019deep} & 5.68  & 24.72 & 3.29 & \textbf{0.78}  & 24.55 & 2.14  & 1.19  & 20.89 & 11.08$\pm$11.61 \\
    \textbf{MCDeepPixBiS-Hyp(ours)} & \textbf{4.46} & 15.30 & 3.70  & 1.64  & 20.56 & \textbf{0.96} & 1.37  & 17.69 & 8.75$\pm$8.69 \\
    \textbf{MCDeepPixBiS-Hyp-HCL(ours)} & 5.08  & \textbf{7.48} & 5.02  & \textbf{0.78} & 19.95 & 4.05  & 0.61 & 16.89 & \textbf{7.83$\pm$7.67} \\
    \bottomrule[2pt]
  \end{tabular}}%
  \label{tab:intra_WMCA}%
\end{table*}%

\begin{table*}[t]
  \caption{Unimodal results of seen and unseen protocols on PADISI-Face dataset. Attacks corresponding to each code are 01-Partial Funny Eyes, 02-Paper, 03-Mask Mannequin, 04-Mask Half, 05-Mask Transparent, 06-Makeup, 07-Mask Silicone, 08-Partial Paper Glasses, 09-Tattoo. The values ACER(\%) reported on testing sets are obtained with thresholds computed for BPCER=1\% on development sets. The best results are bolded.}
\centering
\setlength{\tabcolsep}{11pt}
  \resizebox{16cm}{!}{
     \begin{tabular}{cccccccccccc}
    \toprule[2pt]
    {\textbf{Method}} & {\textbf{Seen}} & \multicolumn{9}{c}{\textbf{Unseen}}                        & {\textbf{Average}} \\
\cmidrule{3-11}          &       & 01 & 02    & 03 & 04 & 05 & 06 & 07 & 08 & 09 &  \\
    \midrule[2pt]
    ViT~\cite{dosovitskiy2020image,liao2023domain}   & 0.28  & 22.33 & 0.27  & 0.54  & 0.54  & 0.27  & 9.49  & 41.18 & 3.99  & 1.80  & 8.93$\pm$14.08 \\
    \textbf{ViT-Hyp (ours)} & \textbf{0.14} & 24.66 & 0.27  & 0.41  & 0.27  & 0.27  & 13.12 & 32.36 & 3.04  & 1.67  & 8.45$\pm$12.23 \\
    \textbf{ViT-Hyp-HCL(ours)} & \textbf{0.14} & 18.04 & 0.27  & 0.27  & \textbf{0.14} & 0.14  & 9.91  & \textbf{10.17} & 5.14  & 2.07  & 5.13$\pm$6.34 \\
   \midrule[2pt]
    MCDeepPixBiS~\cite{george2019deep} & 0.28  & 4.25  & 0.54  & 0.27  & 0.41  & 0.14  & 7.91  & 41.18 & \textbf{0.27} & 0.81  & 6.20$\pm$13.38 \\
    \textbf{MCDeepPixBiS-Hyp(ours)} & \textbf{0.14} & \textbf{1.35} & \textbf{0.14} & \textbf{0.00} & \textbf{0.14} & 0.14  & 10.68 & 37.03 & 0.41  & 0.41  & 5.59$\pm$12.28 \\
    \textbf{MCDeepPixBiS-Hyp-HCL(ours)} & \textbf{0.14} & 3.72  & \textbf{0.14} & 0.41  & \textbf{0.14} & \textbf{0.00} & \textbf{4.43} & 28.89 & \textbf{0.27} & \textbf{0.27} & \textbf{4.25$\pm$9.39} \\
    \bottomrule[2pt]
    \end{tabular}}%
  \label{tab:intra_USC}%
\end{table*}%

\begin{table*}[t]
\caption{Unimodal results of LOO protocols on SiW-M dataset. The values ACER(\%) reported on testing sets are obtained with the threshold of 0.5. The best results are bolded.}
  \centering
   \resizebox{16cm}{!}{
    \begin{tabular}{cccccccccccccccc}
    \toprule[2pt]
\textbf{Method} & \textbf{Metric(\%)} & \textbf{Replay} & {\textbf{Print}} & \multicolumn{5}{c}{\textbf{Mask Attacks}} & \multicolumn{3}{c}{\textbf{Makeup Attacks}} & \multicolumn{3}{c}{\textbf{Partial Attacks}} & \textbf{Average} \\
    \cmidrule(r){5-9}  \cmidrule(r){10-12} \cmidrule(r){13-15} &  &       &       & Half & Silicone & Trans & Paper & Manne & Obfusc & Imperson & Cosmetic & Funny Eye & Paper Glasses & Partial Paper& \\
    \midrule[2pt]
    \multirow{2}{*}{{SVM+LBP}~\cite{zinelabidine2017oulunpu}} & ACER  & 20.6 & 18.4  & 31.3  & 21.4  & 45.5 & 11.6  & 13.8  & 59.3 & 23.9  & 16.7 & 35.9 & 39.2 & 11.7  & 26.9$\pm$14.5 \\
    &EER  & 20.8 & 18.6  & 36.3  & 21.4  & 37.2 & 7.5  & 14.1  & 51.2 & 19.8  & 16.1  & 34.4 & 33.0 & 7.9  & 24.5$\pm$12.9\\
    \hline
    \multirow{2}{*}{{Auxiliary}~\cite{FAS-Auxiliary-CVPR-2018}} & ACER  & 16.8 & 6.9 & 19.3 & 14.9 & 52.1 & 8.0 & 12.8 & 55.8 & 13.7 & 11.7 & 49.0 & 40.5 & 5.3  & 23.6$\pm$18.5 \\
    &EER  & 14.0 & 4.3 & 11.6 & 12.4 & 24.6 & 7.8 & 10.0 & 72.3 & 10.1 & 9.4 & 21.4 & 18.6 & 4.0 & 17.0$\pm$17.7\\
    \hline
    \multirow{2}{*}{{CDCN++}~\cite{yu2020searching}} & ACER  & 10.8 & 7.3 & 9.1 & 10.3 & 18.8 & 3.5 & 5.6 & 42.1 & 0.8 & 14.0 & 24.0 & 17.6 & 1.9 & 12.7$\pm$11.2  \\
    &EER  & 9.2 & 5.6 & 4.2 & 11.1 & 19.3 & 5.9 & 5.0 & 43.5 & \textbf{0.0} & 14.0 & 23.3 & 14.3 & \textbf{0.0} & 11.9$\pm$11.8\\
    \hline
    \multirow{2}{*}{{BCN}~\cite{yu2020face}} & ACER & 12.8 & 5.7 & 10.7 & 10.3 & 14.9 & 1.9 & 2.4 & 32.3 & 0.8 & 12.9 & 22.9 & 16.5 & 1.7 & 11.2$\pm$9.2  \\
    &EER  & 13.4 & 5.2 & 8.3 & 9.7 & 13.6 & 5.8 & 2.5 & 33.8 & \textbf{0.0} & 14.0 & 23.3 & 16.6 & 1.2 & 11.3$\pm$9.5 \\
    \hline
    \multirow{2}{*}{{STDN}~\cite{liu2020disentangling}} & ACER & 7.8 & 7.3 & 7.1 & 12.9 & 13.9 & 4.3 & 6.7 & 53.2 & 4.6 & 19.5 & 20.7 & 21.0 & 5.6 &  14.2$\pm$13.2   \\
    &EER  & \textbf{7.6} & 3.8 & 8.4 & 13.8 & 14.5 & 5.3 & 4.4 & 35.4 & \textbf{0.0} & 19.3 & 21.0 & 20.8 & 1.6 & 12.0$\pm$10.0 \\
    \hline
    \multirow{2}{*}{{DTN}~\cite{DTL}} & ACER & 9.8 & 6.0 & 15.0 & 18.7 & 36.0 & 4.5 & 7.7 & 48.1 & 11.4 & 14.2 & 19.3 & 19.8 & 8.5 & 16.8$\pm$11.1    \\
    &EER & 10.0 & \textbf{2.1} & 14.4 & 18.6 & 26.5 & 5.7 & 9.6 & 50.2 & 10.1 & 13.2 & 19.8 & 20.5 & 8.8 & 16.1$\pm$12.2  \\
    \hline
    \multirow{2}{*}{{DTN(MT)}~\cite{qin2021meta}} & ACER & 9.5 & 7.6 & 13.1 & 16.7 & 20.6 & 2.9 & 5.6 & 34.2 & 3.8 & 12.4 & 19.0 & 20.8 & 3.9 & 13.1$\pm$8.7   \\
    &EER & 9.1 & 7.8 & 14.5 & 14.1 & 18.7 & 3.6 & 6.9 & 35.2 & 3.2 & 11.3 & 18.1 & 17.9 & 3.5 &  12.6$\pm$8.5 \\
    \hline
    \multirow{2}{*}{{FAS-DR(Depth)}~\cite{qin2021meta}} & ACER & 7.8 & 5.9 & 13.4 & 11.7 & 17.4 & 5.4 & 7.4 & 39.0 & 2.3 & 12.6 & 19.6 & 18.4 & 2.4 &  12.6$\pm$9.5  \\
    &EER & 8.0 & 4.9 & 10.8 & 10.2 & 14.3 & 3.9 & 8.6 & 45.8 & 1.0 & 13.3 & 16.1 & 15.6 & 1.2 &  11.8$\pm$11.0 \\
    \hline
    \multirow{2}{*}{{DC-CDN}~\cite{yu2021dual}} & ACER & 12.1 & 9.7 & 14.1 & \textbf{7.2} & 14.8 & 4.5 & \textbf{1.6} & 40.1 & 0.4 & 11.4 & 20.1 & 16.1 & 2.9  &11.9$\pm$10.3 \\
    &EER & 10.3 & 8.7 & 11.1 & 7.4 & 12.5 & 5.9 & \textbf{0.0} & 39.1 & \textbf{0.0} & 12.0 & 18.9 & 13.5 & 1.2 &  10.8$\pm$10.1 \\
    \hline
    \multirow{2}{*}{{FAS-DR(MT)}~\cite{qin2021meta}} & ACER & \textbf{6.3} & \textbf{4.9} & 9.3 & 7.3 & \textbf{12.0} & 3.3 & 3.3 & 39.5 & \textbf{0.2} & 10.4 & 21.0 & 18.4 & \textbf{1.1}  &10.5$\pm$10.3 \\
    &EER & 7.8 & 4.4 & 11.2 & \textbf{5.8} & \textbf{11.2} & 2.8 & 2.7 & 38.9 & 0.2 & 10.1 & 20.5 & 18.9 & 1.3 &  10.4$\pm$10.2 \\
    \midrule[2pt]
    \multirow{2}{*}{{ViT~\cite{dosovitskiy2020image,liao2023domain}}} & ACER  & 11.35 & 5.58  & 3.44  & 9.63  & 16.73 & 1.47  & 2.89  & 26.60 & 1.90  & 9.04  & 23.14 & 11.23 & 2.44  & 9.65$\pm$8.19 \\
    &EER  & 11.18 & 7.32  & 3.89  & 9.63  & 14.32 & 0.00  & 3.50  & 23.48 & 1.64  & 9.20  & 20.38 & 11.32 & 1.86  & 9.06$\pm$7.21\\
    \hline
    \multirow{2}{*}{{MCDeepPixBiS~\cite{george2019deep}}} & ACER  & 9.21  & 8.49  & 4.31  & 9.83  & 15.06 & 3.18  & 2.83  & 48.87 & 2.64  & 16.84 & \textbf{13.71} & 8.38  & 4.03  & 11.34$\pm$12.25 \\
    & EER & 8.90  & 8.19  & 5.00  & 10.37 & 17.05 & \textbf{0.00}  & 3.00  & 29.57 & 2.30  & 14.00 & 17.72 & 9.09  & 4.65  & 9.99$\pm$8.04 \\
    \hline
    \multirow{2}{*}{\textbf{MCDeepPixBiS-Hyp  (ours) }} & ACER  & 9.20  & 6.85  & 5.35  & 8.46  & 14.52 & 3.57  & 1.72  & 40.33 & 2.64  & 12.76 & \textbf{13.71} & \textbf{6.46}  & 3.33  & 9.92$\pm$10.07 \\
     & EER & 10.14 & 8.01  & 6.11  & 8.61  & 18.18 & \textbf{0.00}  & 2.00  & 28.70 & 0.66  & 12.00 & 17.97 & \textbf{7.34}  & 3.02  & 9.44$\pm$8.21 \\
    \hline
    \multirow{2}{*}{\textbf{MCDeepPixBis-Hyp-HCL  (ours)}} & ACER & 10.50 & 7.48  & 4.99  & 7.88  & 16.47 & 2.09  & 3.62  & 20.34 & 3.10  & 14.13 & 14.59 & 8.92  & 2.52  & 8.97$\pm$5.90 \\
    & EER & 10.56 & 7.32  & 5.00  & 9.63  & 19.09 & \textbf{0.00}  & 3.00  & 21.74 & 1.64  & 15.60 & \textbf{15.97} & 8.45  & 3.02  & 9.31$\pm$6.98 \\
     \hline
    \multirow{2}{*}{\textbf{ViT-Hyp (ours)}} & ACER  & 9.90  & 5.52  & 3.46  & 8.47  & 13.46 & 0.98  & 3.08  & 24.26 & 1.78  & 9.49  & 17.71 & 9.50  & 3.80  & 8.57$\pm$6.76 \\
   & EER & 9.94  & 5.40  & 3.89  & 8.15  & 14.32 & 1.18  & 3.50  & 20.87 & 1.31  & 10.00 & 16.33 & 9.50  & 3.95  & \textbf{8.33$\pm$6.02} \\
    \hline
    \multirow{2}{*}{\textbf{ViT-Hyp-HCL (ours)}} & ACER  & 9.72  & 5.57  & \textbf{2.86}  & 9.87  & 15.82 & \textbf{0.08}  & 2.63  & \textbf{18.02} & 0.94  & \textbf{8.30}  & 20.52 & 12.02 & 3.45  & \textbf{8.45$\pm$6.66} \\
     & EER & 9.94  & 5.75  & \textbf{3.06}  & 9.63  & 13.86 & \textbf{0.00} & 2.50  & \textbf{18.26} & 0.98  & \textbf{8.00}  & 23.92 & 11.64 & 4.19  & 8.59$\pm$7.05 \\
    \bottomrule[2pt]
    \end{tabular}}
  \label{tab:intra_SiWM}%
\end{table*}%

\subsection{Implementation Details}
\label{sec:Details}
Two favorite backbones (i.e., MCDeepPixBiS \cite{george2019deep} and Vision Transformer (ViT) \cite{dosovitskiy2020image,liao2023domain}) are utilized in our experiments. Specifically, we train all layers of MCDeepPixBiS-variant models. For the ImageNet pretrained ViT backbone, we only train its last 8 encoder layers. To ease the demonstration of different backbones and their variants in unimodal learning, we define the below notations:
 \begin{itemize}
\setlength\itemsep{-0.1em}
     \item \textbf{Backbone}: The vanilla backbone models (MCDeepPixBiS/ ViT) optimized by standard cross entropy loss; 
   \item  \textbf{Backbone-Hyp}: The backbone in our proposed hyperbolic framework, but replaces standard cross entropy loss with $\mathcal{L}^{Hyp-BCE}$ for optimization;
   \item  \textbf{Backbone-Hyp-HCL}: On the basis of the Backbone-Hyp model, $\mathcal{L}^{BF}$ is added for optimization.
\end{itemize}

Similarly, for multimodal learning, we define the below notations:
 \begin{itemize}
\setlength\itemsep{-0.1em}
     \item \textbf{Backbone-Euc-Euc}: Unimodal features extracted by the backbone are optimized by Euclidean multimodal feature decomposition and Euclidean multimodal feature fusion \& classification. Hyperbolic contrastive loss in the proposed framework is replaced by Euclidean contrastive loss; 
     \item \textbf{Backbone-Euc-Hyp}: Unimodal features extracted by the backbone are optimized by Euclidean multimodal feature decomposition and hyperbolic multimodal feature fusion \& classification.
\end{itemize}

It is worth noting that in the MCDeepPixBiS model, in addition to standard cross entropy loss, a pixel-wise binary loss $\mathcal{L}_{pixel-wise}$ is also used for optimization. For the WMCA dataset, 5 frames were randomly selected from each video. For the SiW-M dataset, in order to exclude the influence of the background in the image on face detection, we use MTCNN \cite{zhang2016joint} to crop the face regions in the image.

By default, in our hyperbolic framework, we set the hyperbolic curvature $c=0.1$, the ineffective radius proportions $\alpha = 0.1$, and the hyperbolic embedding dimension as 128 for the unimodal intra-dataset testing. For the loss function of ViT-Hyp-HCL model, we use $\lambda_1 = 1.0$, $\lambda_2 = 0.1$  and $\tau = 1.0$. For the loss function of the MCDeepPixBiS-Hyp-HCL model, $\lambda_1$ is changed to 0.5, and the coefficient of $\mathcal{L}_{pixel-wise}$ is 0.5. During the training of the ViT-Euc-Hyp/MCDeepPixBiS-Euc-Hyp model, $\gamma_1 = \gamma_2 =\gamma_3 = \gamma_4 = 1.0$ on the WMCA dataset. The coefficient of $\mathcal{L}_{pixel-wise}$ is set to 0.1 in the MCDeepPixBiS-Euc-Hyp model. $\gamma_1$, $\gamma_3$, $\gamma_4$ are changed to 0.2, and $c$ is set to 0.5 for the ViT-Euc-Hyp model training on the PADISI-Face dataset. For the loss function of the MCDeepPixBiS-Euc-Hyp model training on the PADISI-Face dataset, $\gamma_1=0.05$, $\gamma_3 = \gamma_4 =0.01$, and the coefficient of $\mathcal{L}_{pixel-wise}$ is 0.5. 

As for data pre-processing, face images in RGB modality are resized to $224\times224\times3$ and face images in Depth/IR modality are resized to $224\times224$. To increase model robustness, for RGB images, data augmentations including horizontal flipping with a probability of 0.5, adding a random value $\Delta \in [-40,40]$ to all pixels per image channel, and random $\gamma$-correction with a random $\gamma\in[0.5,1.5]$ are used. During the training process, Adam Optimizer with a weight decay parameter of $1\times 10^{-5}$ is utilized. The batch size is 64 for unimodal training and 32 for multimodal training, and the number of total training epochs is 30. The initial learning rate is $1\times 10^{-4}$ and the learning rate is multiplied by 0.8 after every 10 epochs.

\subsection{Unimodal Intra-dataset Testing}
\label{sec:Intra_Testing}
\textbf{Results on WMCA dataset.} The experimental results of the grandtest (seen) and LOO unseen attack protocols of WMCA are shown in Table~\ref{tab:intra_WMCA}. We can see that the ACER of the ViT-Hyp-HCL model on the seen protocol is 5.62\%, which is 1.27\% lower than the baseline ViT model. The effect is more significant on unseen protocols as ACER decreased from 14.83$\pm$12.00\% to 11.52$\pm$10.68\%. Moreover, we can observe the MCDeepPixBiS-Hyp and MCDeepPixBiS-Hyp-HCL surpass the baseline MCDeepPixBis by a clear margin. Therefore, the proposed hyperbolic unimodal framework and hyperbolic contrastive loss can significantly improve generalization performance against unseen attacks.

\begin{table*}[t]
  \caption{Unimodal results of one-to-one cross-dataset testings among WMCA, PADISI-Face, and SiW-M datasets. HTER(\%) values reported on testing sets are obtained with the fixed threshold of 0.5 on testing sets. The best results are bolded.}
\belowrulesep=0ex
\aboverulesep=0ex
\centering
\setlength{\tabcolsep}{11pt}
  \resizebox{1.0\linewidth}{!}{
\begin{tabular}{|c|c|c|c|c|c|c|c|c|c|c|c|c|c|c|}
    \hline 
    \textbf{Training Dataset} & \multicolumn{4}{c|}{WMCA} & \multicolumn{4}{c|}{PADISI} & \multicolumn{4}{c|}{SiW-M}& \multicolumn{2}{c|}{Average}\\
    \cmidrule{1-13}\textbf{Testing Dataset}  & \multicolumn{2}{c|}{PADISI} & \multicolumn{2}{c|}{SiW-M} & \multicolumn{2}{c|}{WMCA} & \multicolumn{2}{c|}{SiW-M} & \multicolumn{2}{c|}{WMCA} & \multicolumn{2}{c|}{PADISI}&\multicolumn{2}{c|}{}\\
    \hline
    \textbf{Metric} & AUC(\%) & HTER(\%) & AUC(\%) & HTER(\%)  & AUC(\%) & HTER(\%) & AUC(\%) & HTE(\%)R & AUC(\%) & HTER(\%) & AUC(\%) & HTER(\%) & AUC(\%) & HTER(\%) \\
    \hline
    ViT~\cite{dosovitskiy2020image,liao2023domain}  &97.30 & 10.06 & 86.77 & 28.03 &92.49 & 16.26 & 85.82 & 24.70 & 90.22 & 18.46 &  98.85 & 5.65& 91.91$\pm$5.36 & 17.19$\pm$8.49 \\
    \textbf{ViT-Hyp(ours)}& 97.36 & 9.63  & 87.02 & 26.66 & 93.52 & 15.43 & 85.26 & 23.56 &  89.84 & 21.21 & 98.82 & 5.57& 91.97$\pm$5.52& 17.01$\pm$8.26 \\
     \textbf{ViT-Hyp-HCL(ours)}& \textbf{97.80} & \textbf{8.98}  & 87.99 & 23.33 & \textbf{94.80} & \textbf{12.07} & 86.80 & 20.91 & \textbf{91.34} & \textbf{16.87} &  \textbf{99.18} & \textbf{5.02}& \textbf{92.99}$\pm$\textbf{5.11} & \textbf{14.53}$\pm$\textbf{7.08} \\
    \hline
    MCDeepPixBiS~\cite{george2019deep} & 95.43 & 12.83 & \textbf{89.47} & 22.67 & 90.96 & 17.18 & 86.23 & 26.02 & 88.01 & 23.64 &  97.00 & 9.61 & 91.18$\pm$4.23 & 18.66$\pm$6.53 \\
    \textbf{MCDeepPixBiS-Hyp(ours)} & 93.98 & 13.02 & 89.14 & 26.87 & 92.79 & 17.67 & \textbf{88.97} & \textbf{19.12} & 87.61 & 22.36 &  96.52 & 9.36 & 91.50$\pm$3.47 & 18.07$\pm$6.30 \\
    \textbf{MCDeepPixBiS-Hyp-HCL(ours)} & 95.09 & 11.00 & 88.61 & \textbf{20.57} &93.25 & 16.55 & 87.58 & 21.37 & 90.63 & 20.21 &  98.44 & 7.08 &  92.27$\pm$4.12 & 16.13$\pm$5.87  \\
    \hline
\end{tabular}}%
  \label{tab:cross_test_11}%
\end{table*}%

\begin{table*}[t]
  \caption{Unimodal results of two-to-one cross-dataset testings among WMCA, PADISI-Face, and SiW-M datasets. HTER(\%) values reported on testing sets are obtained with the fixed threshold of 0.5 on testing sets. The best results are bolded.}
\belowrulesep=0ex
\aboverulesep=0ex
\centering
\setlength{\tabcolsep}{11pt}
  \resizebox{1.0\linewidth}{!}{
\begin{tabular}{|c|c|c|c|c|c|c|c|c|}
    \hline
    \textbf{Method} & \multicolumn{2}{c|}{\textbf{WMCA \& PADISI to SiW-M}} & \multicolumn{2}{c|}{\textbf{WMCA \& SiW-M to PADISI}} & \multicolumn{2}{c|}{\textbf{PADISI \& SiW-M to WMCA}}& \multicolumn{2}{c|}{\textbf{Average}}\\
    \cmidrule{2-9} & AUC (\%) & HTER (\%) & AUC (\%) & HTER (\%)  & AUC (\%) & HTER (\%) & AUC (\%) & HTER (\%) \\
    \hline
    ViT~\cite{dosovitskiy2020image,liao2023domain}   & 89.43  & 21.99 & 99.20  & 4.10 & 93.69 & 14.86 & 94.11$\pm$4.90 & 13.65$\pm$9.01 \\
    \textbf{ViT-Hyp (ours)} & 88.94 & 21.63& 99.35 & 4.16 & \textbf{94.44} & \textbf{13.99} & 94.24$\pm$5.21 & 13.26$\pm$8.76 \\
    \textbf{ViT-Hyp-HCL(ours)} & 89.08 & 18.15& \textbf{99.47} & \textbf{3.59} &94.19 &14.66 &\textbf{94.25}$\pm$\textbf{5.20} & \textbf{12.13}$\pm$\textbf{7.60} \\
    \hline
    MCDeepPixBiS~\cite{george2019deep} & 91.38 & 19.61 & 98.49 &6.52 & 90.91 & 18.86 & 93.59$\pm$4.25 & 15.00$\pm$7.35 \\
    \textbf{MCDeepPixBiS-Hyp(ours)} & \textbf{91.71} &20.87 & 98.64 &4.99 & 91.66 &16.48 & 94.00$\pm$4.02 & 14.11$\pm$8.20 \\
    \textbf{MCDeepPixBiS-Hyp-HCL(ours)} & 91.21 & \textbf{16.44} & 98.91 &  5.16 & 92.07 & 16.47 &94.06$\pm$4.22 & 12.69$\pm$6.52  \\
    \hline
\end{tabular}}%
  \label{tab:cross_test_21}%
\end{table*}%

\begin{table*}[t]
  \caption{Mutimodal results of seen and unseen protocols on WMCA dataset. The values ACER(\%) reported on testing sets are obtained with thresholds computed for BPCER=1\% on development sets. The best results are bolded.}
\centering
\setlength{\tabcolsep}{11pt}
  \resizebox{16cm}{!}{
        \begin{tabular}{ccccccccccc}
    \toprule[2pt]
    {\textbf{Modality}}&{\textbf{Method}} & {\textbf{Seen}} & \multicolumn{7}{c}{\textbf{Unseen}}                  & {\textbf{Average}} \\
\cmidrule{4-10}    &      &       & Glasses & Rigid Mask & Fake Head & Flexible Mask & Paper Mask & Print & Replay &  \\
    \midrule[2pt]
    \multirow{7}{*}
     &MCCNN-OCCL-GMM~\cite{george2020learning} &  3.30& 50.00 & 18.30 & 1.90  & 22.80 & 4.80 & 30.00  & 31.40 & 22.74$\pm$15.30 \\{\multirow{8}{*}{\textbf{RGB+Depth}}}&MCDeepPixBiS~\cite{george2019deep} &  1.80 & \textbf{16.00} & 3.40 & 0.70  & 49.70 & 0.20 & 0.10  & 3.70  & 10.50$\pm$16.70 \\
    &CMFL~\cite{george2021cross} &  1.70& 33.50 & 1.70 & 2.50  & \textbf{12.40} & 1.80 & 0.70  & 1.00 & 7.60$\pm$11.20 \\
    &ViT~\cite{dosovitskiy2020image,liao2023domain} &  3.79& 36.00& 2.19 & 4.14  & 14.56 & 2.82 & 0.83  & 1.58 & 7.93$\pm$13.80 \\
    &\textbf{ViT-Euc-Euc (ours)} & 1.62  & 25.60 & 0.78  & 1.30  & 18.42 & \textbf{0.09}  & 0.24  & 0.43 & 6.69$\pm$10.67 \\
    &\textbf{ViT-Euc-Hyp(ours)} & 1.65  & 26.03 & 0.70  & \textbf{0.26}  & 17.86 & \textbf{0.09}  & \textbf{0.00}  & \textbf{0.00} & 6.42$\pm$10.87 \\
    &\textbf{MCDeepPixBiS-Euc-Euc(ours)} & 0.57 & 28.18 & \textbf{0.00}  & 4.40  & 16.03 & 0.17 & \textbf{0.00}  & 1.61 & 7.20$\pm$10.88 \\
    &\textbf{MCDeepPixBiS-Euc-Hyp(ours)}  & \textbf{0.50} & 16.72 & 0.21  & 1.39  & 18.49 & \textbf{0.09} & 0.24  & 1.30 & \textbf{5.49$\pm$8.31} \\
    \midrule[2pt]
    \multirow{4}{*}{\textbf{RGB+IR}}&\textbf{ViT-Euc-Euc (ours)} & 2.49  & 33.72 & 1.53  & 11.13 & 10.31 & \textbf{0.00}  & \textbf{0.00}  & \textbf{0.00} & 8.10$\pm$12.32\\
    &\textbf{ViT-Euc-Hyp (ours)} & 2.41  & 31.36 & 0.99  & 3.87 & 9.74 & \textbf{0.00}  & \textbf{0.00}  & \textbf{0.00} & 6.57$\pm$11.49\\
     &\textbf{MCDeepPixBiS-Euc-Euc(ours)} & 0.45 & 24.55 & \textbf{0.00}  & 0.35  & 5.21 & 0.09 & \textbf{0.00}  & 0.17 & 4.34$\pm$9.11 \\
    &\textbf{MCDeepPixBiS-Euc-Hyp(ours)}  & \textbf{0.25} & \textbf{15.62} & 0.26  & \textbf{0.00}  & \textbf{3.09} & 0.09 &  \textbf{0.00} & \textbf{0.00} & \textbf{2.72$\pm$5.80} \\
    \bottomrule[2pt]
  \end{tabular}}%
  \label{tab:intra_WMCA_MM}%
\end{table*}%

\textbf{Results on PADISI-Face dataset.} The results of the PADISI-Face dataset are shown in Table \ref{tab:intra_USC}. While all the models perform comparatively with ACER lower than 0.3\% on the seen protocol, our hyperbolic models show clear advantages over the baseline models on the more challenging unseen protocol. Specifically, our ViT-Hyp-HCL model reduces the ACER from 8.93$\pm$14.08\% to 5.13$\pm$6.34\%, and our MCDeepPixBiS-Hyp-HCL model reduces the ACER from 6.20$\pm$13.38\% to 4.25 $\pm$9.39\%. Therefore, the proposed hyperbolic framework is beneficial on different baseline models (e.g., ViT and MCDeepPixBiS) against unseen attack types.

\textbf{Results on SiW-M dataset.} Experimental results of intra-dataset testing on the SiW-M dataset are shown in \autoref{tab:intra_SiWM}. It can be seen that compared with ViT/MCDeepPixBiS baseline models, hyperbolic layers and hyperbolic contrastive loss can bring significant improvements. Especially for the MCDeepPixBiS-Hyp-HCL model, it can decrease the average ACER from 11.34$\pm$12.25\% to 8.97$\pm$5.90\%, and the average EER from 9.99$\pm$8.04\% to 9.31$\pm$6.98\% respectively. The other important observation is that the proposed models can have better performance than other state-of-the-art methods. ViT-Hyp achieves the best average EER of 8.33$\pm$6.02\% and ViT-Hyp-HCL achieves the best average ACER of 8.45$\pm$6.66\%. Hence, the hyperbolic frameworks can boost models' generalization ability against unknown attacks on the SiW-M dataset.

\subsection{Unimodal Cross-dataset Testing} \label{sec:Inter_Testing}
Here we evaluate our methods under cross-domain attack-agnostic scenarios by conducting cross-dataset testings among WMCA, PADISI-Face, and SiW-M datasets.

\textbf{Results on one-to-one cross-dataset testing.} The experimental results of training one dataset and testing on the other dataset are shown in \autoref{tab:cross_test_11}. The highlight is that the proposed ViT-Hyp-HCL can significantly reduce average HTER from 17.19$\pm$8.49\% of the baseline to 14.53$\pm$7.08\%. The MCDeepPixBiS-Hyp-HCL can also improve the average HTER from 18.66$\pm$6.53\% to 16.13$\pm$5.87\%.

\begin{figure}[t]
\centering
\includegraphics[width=1.0\linewidth]{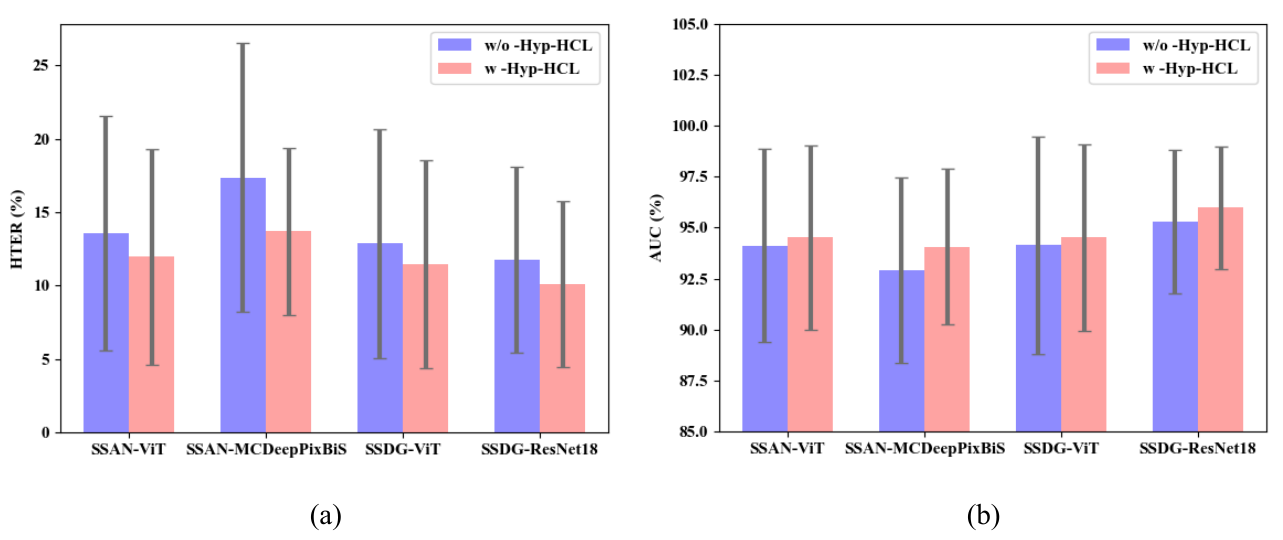}
\caption{Different domain generalization models in terms of (a) HTER; (b) AUC.}
\label{fig:DG_models}
\end{figure}

\textbf{Results on two-to-one cross-dataset testing.} The experimental results of training on two datasets and testing on the other remaining dataset are shown in \autoref{tab:cross_test_21}. The results illustrate that both ViT-Hyp-HCL and MCDeepPixBiS-Hyp-HCL can achieve impressive improvements in average HTER compared with baseline models. For example, MCDeepPixBiS-Hyp-HCL reduces the HTER from 15.00$\pm$7.35\%  to 12.69$\pm$6.52\%. Furthermore, to verify the effectiveness of our methods on classical domain generalization FAS frameworks, we implement SSAN~\cite{wang2022domain} and SSDG~\cite{jia2020single} for cross-dataset testings combined with the proposed hyperbolic framework and different backbones. The experimental results are shown in \autoref{fig:DG_models}. Same as the protocols in \autoref{tab:cross_test_21}, these models are also trained on two datasets of WMCA/PADISI-Face/SiW-M and tested on the remaining dataset. The average AUC and HTER are reported. It can be seen that the proposed hyperbolic framework can improve model performance in these domain generalization models.

Hence, it can be concluded that our proposed hyperbolic unimodal FAS framework can elevate the generalization capacity of FAS models against agnostic unseen attacks in unseen domains.

\subsection{Multimodal Intra-dataset Testing}
\label{sec:MM_Intra_Testing}

\begin{table*}[t]
  \caption{Multimodal results of seen and unseen protocols on PADISI-Face dataset. The values ACER(\%) reported on testing sets are obtained with thresholds computed for BPCER=1\% on development sets. The best results are bolded.}
\centering
\setlength{\tabcolsep}{11pt}
  \resizebox{16cm}{!}{
     \begin{tabular}{ccccccccccccc}
    \toprule[2pt]
    {\textbf{Modality}}& {\textbf{Method}} & {\textbf{Seen}} & \multicolumn{9}{c}{\textbf{Unseen}}                        & {\textbf{Average}} \\
\cmidrule{4-12}    &      &       & 01 & 02    & 03 & 04 & 05 & 06 & 07 & 08 & 09 &  \\
    \midrule[2pt]
    \multirow{4}{*}{\textbf{RGB+Depth}} &\textbf{ViT-Euc-Euc (ours)} & 0.55 & 24.43 & 0.27  & \textbf{0.27}  & 0.41  & \textbf{0.14}  & 16.20 & 14.31 & 3.18  & 0.41  & 6.62$\pm$9.22 \\
     &\textbf{ViT-Euc-Hyp(ours)} & \textbf{0.14} & 20.54 & \textbf{0.00}  & 0.41  & 0.41 & \textbf{0.14}  & 8.37  & \textbf{9.50} & 5.14  & \textbf{0.27}  & 4.98$\pm$6.95 \\
    &\textbf{MCDeepPixBiS-Euc-Euc(ours)} & 0.33 & 4.53 & 0.54& 0.54 & \textbf{0.14} & 0.41  & \textbf{5.43} & 32.63 & \textbf{0.41}  & 0.95  & 5.06$\pm$10.52 \\
    &\textbf{MCDeepPixBiS-Euc-Hyp(ours)} & \textbf{0.14} & \textbf{1.35}  & 0.14 & \textbf{0.27}  & 0.95 & 0.27 & 5.70 & 18.45 & \textbf{0.41} & 0.81 & \textbf{3.15$\pm$5.99} \\
    \midrule[2pt]
    \multirow{4}{*}{\textbf{RGB+IR}} &\textbf{ViT-Euc-Euc (ours)} & 0.30 & 22.74 & \textbf{0.00}  & 0.41  & \textbf{0.27}  & 0.27  & 13.12 & 23.27 & 2.91  & \textbf{0.27}  & 7.03$\pm$9.97 \\
     &\textbf{ViT-Euc-Hyp(ours)} & 0.28 & 22.23 & \textbf{0.00}  & 0.41  & \textbf{0.27} & \textbf{0.00}  & 11.17 & \textbf{5.49} & 3.31  & 0.68  & 4.84$\pm$7.50 \\
    &\textbf{MCDeepPixBiS-Euc-Euc(ours)} & 0.28 & 3.72 & 0.41 & 0.54 & 0.54 & 0.64  & \textbf{3.75} & 33.44 & 0.54  & 2.07  & 5.07$\pm$10.73 \\
    &\textbf{MCDeepPixBiS-Euc-Hyp(ours)} & \textbf{0.00} & \textbf{3.01}  & 0.14 & \textbf{0.27} & 0.68 & 0.41 & 6.96 & 23.13 & \textbf{0.41} & 2.21 & \textbf{4.14$\pm$7.45} \\
    \bottomrule[2pt]
    \end{tabular}}%
  \label{tab:intra_USC_MM}%
\end{table*}%

\textbf{Results on WMCA dataset.} The multimodal experimental result of the seen and LOO unseen attack protocols of WMCA are shown in \autoref{tab:intra_WMCA_MM}. It can be seen that the proposed hyperbolic multimodal FAS framework is effective for classification under RGB + Depth/IR modalities. The proposed framework can bring significant improvement compared with several state-of-art methods in seen and unseen scenarios. And it is worth noting that hyperbolic multimodal feature fusion \& classification can achieve better results than Euclidean multimodal feature fusion \& classification. For the RGB+Depth modalities, MCDeepPixBiS-Euc-Hyp achieves the best performance in the seen task with an ACER of 0.50\% and unseen tasks with an average ACER of 5.49$\pm$8.31\%. It also performs best in the RGB+IR modalities, especially on the unseen tasks with an average ACER of 2.72$\pm$5.80\%. Therefore, it can be concluded that this proposed multimodal framework is beneficial for the model to learn multimodal features.

\textbf{Results on PADISI-Face dataset.} 
Multimodal results of the PADISI-Face dataset are shown in \autoref{tab:intra_USC_MM}. All the models perform comparatively with ACER lower than 0.55\% on the seen protocol. It can be seen that the hyperbolic multimodal feature fusion \& classification can boost the model performance on the unseen tasks compared with Euclidean baselines. The most significant improvement is achieved by the ViT-Euc-Hyp model which decreases the average ACER from 7.03$\pm$9.97\% from 4.84$\pm$7.50\%. The lowest average ACER on unseen protocols is 3.15$\pm$5.99\% for RGB+Depth modalities and 4.14$\pm$7.45\% for RGB+IR modalities, which are both obtained from MCDeepPixBiS-Euc-Hyp model. Hence, the hyperbolic multimodal feature fusion \& classification can improve the generalization ability of models.

\begin{table*}[t]
  \caption{Mutimodal results of one-to-one cross-dataset testings between WMCA and PADISI-Face datasets. HTER(\%) values reported on testing sets are obtained with the fixed threshold of 0.5 on testing sets. The best results are bolded.}
\belowrulesep=0ex
\aboverulesep=0ex
\centering
\setlength{\tabcolsep}{11pt}
  \resizebox{1.0\linewidth}{!}{
\begin{tabular}{|c|c|c|c|c|c|c|c|c|c|c|}
    \hline
    \textbf{Method} & \multicolumn{4}{c|}{\textbf{WMCA to PADISI}} & \multicolumn{4}{c|}{\textbf{PADISI to WMCA}}& \multicolumn{2}{c|}{\multirow{2}{*}{\textbf{Average}}}\\
    \cmidrule{2-9} & \multicolumn{2}{c|}{RGB+Depth} &  \multicolumn{2}{c|}{RGB+IR}&\multicolumn{2}{c|}{RGB+Depth} &  \multicolumn{2}{c|}{RGB+IR}&\multicolumn{2}{c|}{} \\
    \cmidrule{2-11} & AUC (\%) & HTER (\%)& AUC (\%) & HTER (\%)& AUC (\%) & HTER (\%) & AUC (\%) & HTER (\%)  & AUC (\%) & HTER (\%)\\
    \hline
    \textbf{ViT-Euc-Euc (ours)}& 96.28&11.78&95.97&11.90&94.19&13.14&93.15&15.39&94.90$\pm$1.49&13.05$\pm$1.68\\
    \textbf{ViT-Euc-Hyp (ours)}& \textbf{96.30}&\textbf{11.29}&\textbf{97.02}&\textbf{8.50}&\textbf{95.04}&\textbf{12.55}&\textbf{96.04}&\textbf{12.0}5&\textbf{96.10$\pm$0.82}&\textbf{11.10$\pm$1.81}\\
    \textbf{MCDeepPixBiS-Euc-Euc (ours)}& 94.80&20.06&90.52&16.68&94.30&15.90&92.53&15.51&93.04$\pm$1.94&17.04$\pm$2.07\\
    \textbf{MCDeepPixBiS-Euc-Hyp (ours)}& 94.89&13.82&92.46&15.37&93.97&15.63&93.86&14.20&93.80$\pm$1.00&14.76$\pm$0.88\\
    \hline
\end{tabular}}%
  \label{tab:cross_test_MM}%
\end{table*}%

\subsection{Mutimodal Cross-dataset Testing}
Results of multimodal cross-dataset testings between WMCA and PADISI-Face datasets are shown in \autoref{tab:cross_test_MM}. It illustrates that the proposed framework with hyperbolic multimodal feature fusion \& classification can achieve better AUC(\%) and HTER(\%) than Euclidean multimodal feature fusion \& classification. ViT-Euc-Hyp model can increase AUC(\%) from 94.90$\pm$1.49\% to 96.10$\pm$0.82\%. And MCDeepPixBiS-Euc-Hyp decreases HTER(\%) from 17.04$\pm$2.07\% to 14.76$\pm$0.88\%. Therefore, the proposed hyperbolic multimodal feature fusion \& classification can boost the model's generalization ability against unseen domains. However, compared with unimodal results in \autoref{tab:cross_test_11}, the multimodal FAS framework fails to bring significant improvements in the cross-dataset protocols between WMCA and PADISI-Face datasets. This may be because training on two modalities results in larger domain shifts between WMCA and PADISI-Face datasets, which brings challenges to model generalization.

\subsection{Ablation Study}
\label{sec:Ablation}

\begin{figure*}[t]
\centering
\includegraphics[width=16cm]{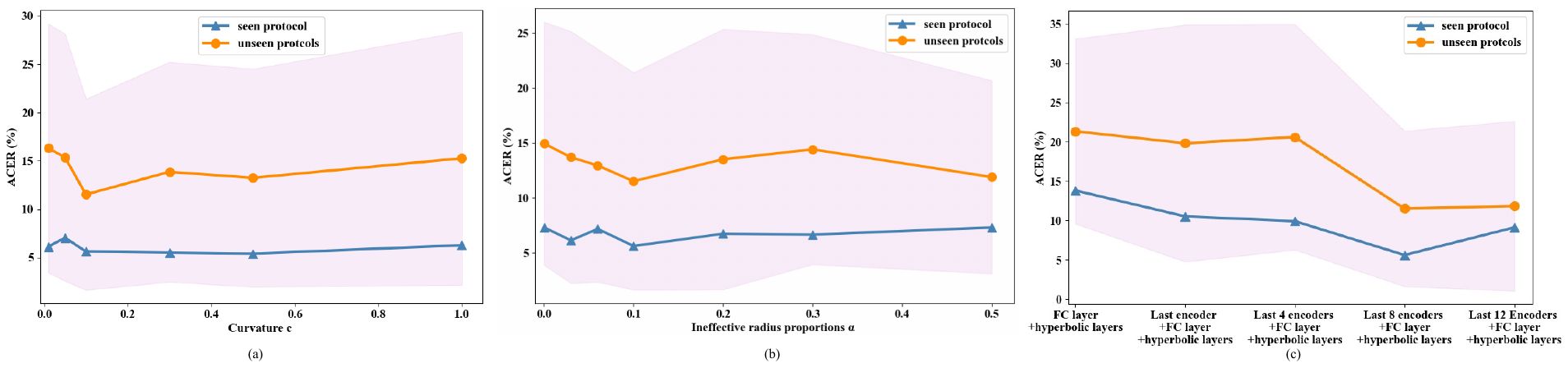}
\caption{The performance of the ViT-Hyp-HCL model under different (a) curvature $c$; (b) ineffective radius proportions $\alpha$; (c) tuned layers. The orange lines denote the mean ACER(\%) value while the plum areas denote the range of standard deviation.}

\label{fig:abl_line}
\end{figure*}%
\textbf{Impact of curvature $c$ of the Poincar\'e ball.} The curvature $c$ of the Poincar\'e ball can be seen as a trade-off between Euclidean space and hyperbolic space. When the curvature $c$ approaches zero, the embedding space is closer to Euclidean space. Conversely, the larger the curvature $c$, the smaller the radius of the Poincar\'e ball, which means we have a smaller hyperbolic space to embed data. In our experiments, the curvature $c$ is set to 0.1. It can be found from \autoref{fig:abl_line}(a) that when the curvature $c = 0.1$, the ViT-Hyp-CL model performs the best in the unseen protocols. In the seen protocol, when the curvature $c$ is 0.3 or 0.5, comparable results are also achieved, but no significant improvement. Since the real-world deployment is closer to the scenario simulated by unseen protocols, we choose $c = 0.1$ as our experimental parameter.

\textbf{Impact of the effective radius of the Poincar\'e ball.}
As discussed in Section \ref{sec:hyp_clip}, in order to avoid the vanishing gradient problem, we propose to limit the maximum norm of the point on the Poincar\'e ball, that is, to specify the effective radius of the Poincar\'e ball. The influence of hyperparameters $\alpha$ on the ViT-Hyp model is shown in Figure \ref{fig:abl_line}(b). When the effective radius of the Poincar\'e ball is small, the embeddable hyperbolic space will be compressed. When the effective radius is large, the resulting vanishing gradient problem will make training difficult. Therefore, we need to balance the embedding space capacity and the difficulty of training. Figure \ref{fig:abl_line}(b) shows that when $\alpha =0.1$, our ViT-Hyp-CL model performs best on the seen protocol and unseen protocols. Hence, $\alpha =0.1$ is assigned in our experiment.

\textbf{Impact of tuned layers in ViT model.} Because we use the ViT model pretrained on ImageNet as the backbone, the effect of freezing some model layers during training can be explored. In Figure \ref{fig:abl_line}(c), we conduct ablation experiments on different tuned layers. When the number of tuned layers is small, the flexibility of the model will be insufficient, which will make the model more difficult to learn. When the number of tuned layers is large, since there are not a large number of images in the WMCA dataset, the model is prone to overfitting. Figure \ref{fig:abl_line}(c) shows that the ViT-Hyp-HCL model works best on the default configuration.

\begin{figure}[t]
\centering
\includegraphics[width=1.0\linewidth]{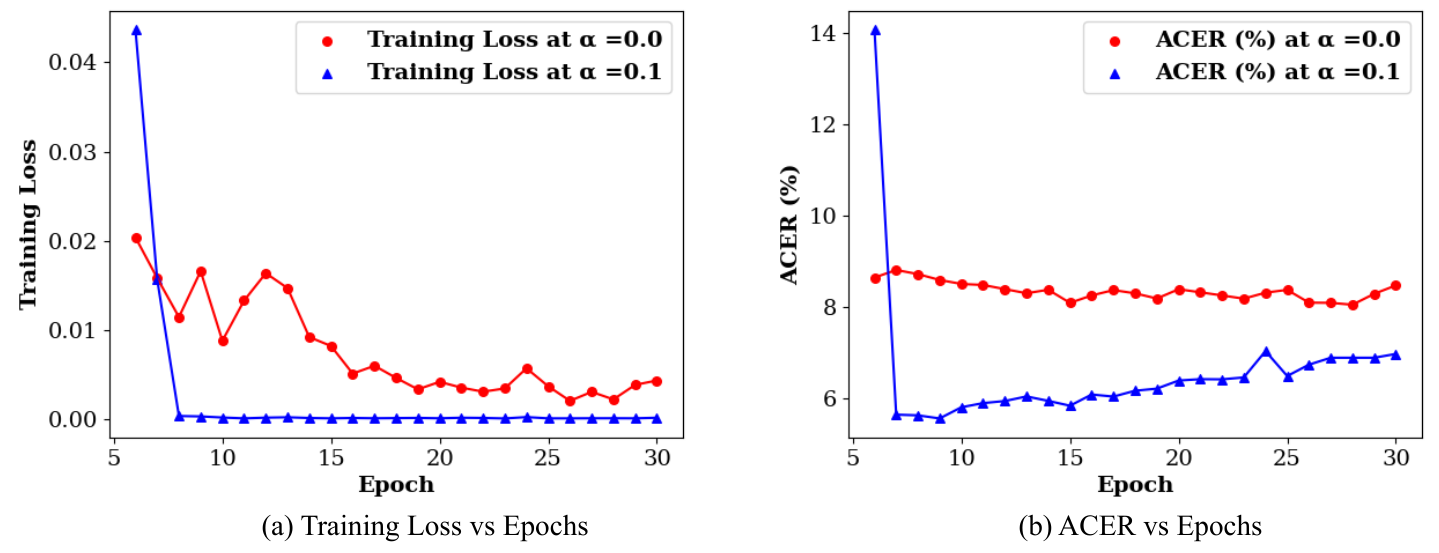}
\caption{Training loss and ACER against epoch at $\alpha$ = 0.1 and $\alpha$ = 0.0. The first 5 epochs are discarded for better visual effects.}
\label{fig:Visual_clip}
\end{figure}%

\textbf{Effect of the proposed hyperbolic feature clipping.} In order to verify the discussion about the vanishing gradient problem in Section \ref{sec:hyp_clip}, we plotted the training loss and ACER of the testing set with epoch at $\alpha =0$ and $\alpha = 0.1$. The results are shown in Figure \ref{fig:Visual_clip}. It can be found that the training loss at $\alpha =0$ is more difficult to converge than the training loss at $\alpha =0.1$, which is caused by the vanishing gradient problem. The vanishing gradient problem will result in the neural network parameters being hard to optimize.

\begin{figure}[t]
\centering
\includegraphics[width=1.0\linewidth]{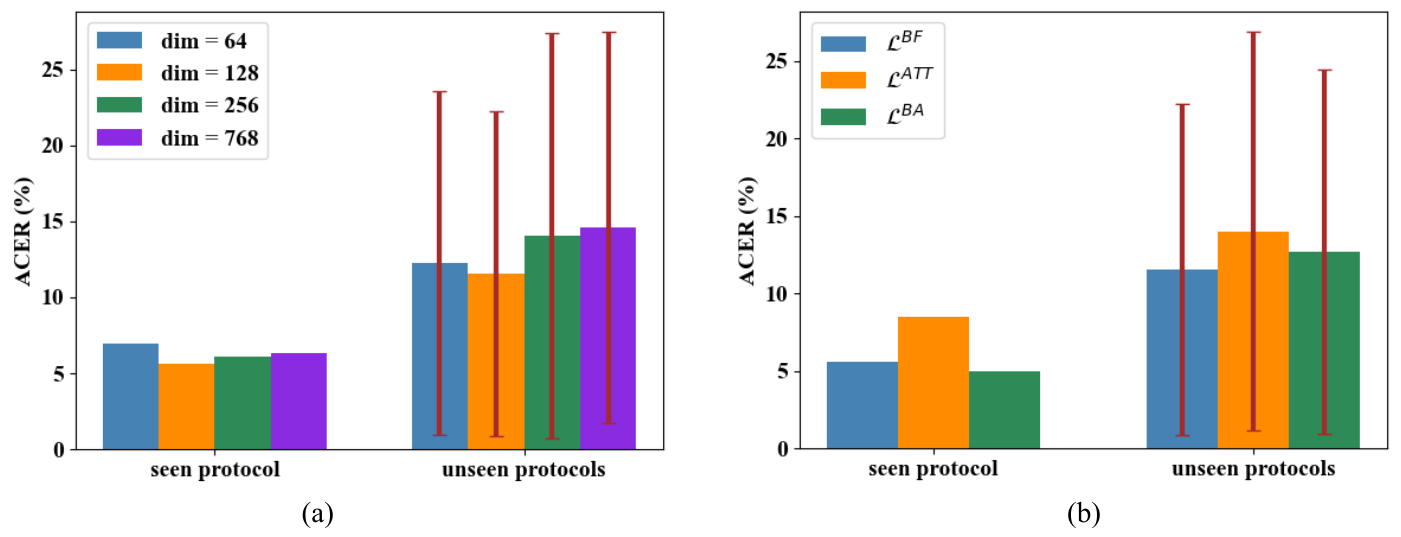}
\caption{Results of the ViT-Hyp-HCL models under different (a) hyperbolic mapping dimensions; (b) hyperbolic contrastive losses.}
\label{fig:abl_bar}
\end{figure}%

\textbf{Impact of the hyperbolic embedding dimension.} The hyperbolic embedding dimension refers to the dimension of feature vectors before the classification is completed in hyperbolic space. Generally speaking, the larger the dimension of the feature embeddings, the richer the features extracted by the model. But this would raise the risk of redundancy and overfitting issues. Some extracted features may only be specific to the training set rather than universal, which will reduce the generalization ability of the model. Therefore, a suitable hyperbolic embedding dimension benefits efficient and generalized feature representation. It can be seen from Figure \ref{fig:abl_bar}(a) that the ViT-Hyp-HCL model achieves the best performance on both seen and unseen protocols when the hyperbolic embedding dimension is 128, which is used as defaulted dimension setting.

\textbf{Impact of the hyperbolic contrastive loss patterns.} Figure \ref{fig:abl_bar}(b) shows the comparison of model effects under different hyperbolic contrastive loss patterns which are discussed in section \ref{sec:hyp_contra}. It can be seen that $\mathcal{L}^{ATT}$  will bring negative effects to the training of the model. It is worth noting that, when $\mathcal{L}^{BA}$ is adopted as our hyperbolic contrastive loss, the model performs the best on the seen protocol. We argue that this is because learning to shrink the distance between attack samples improves the model's detection performance for seen attacks but at the expense of weakened generalization ability. In our experiment, we put the generalization ability of the model first when evaluating the model, hence  $\mathcal{L}^{BF}$ is chosen as the hyperbolic contrastive loss.

\begin{figure}[t]
\centering
\includegraphics[width=1.0\linewidth]{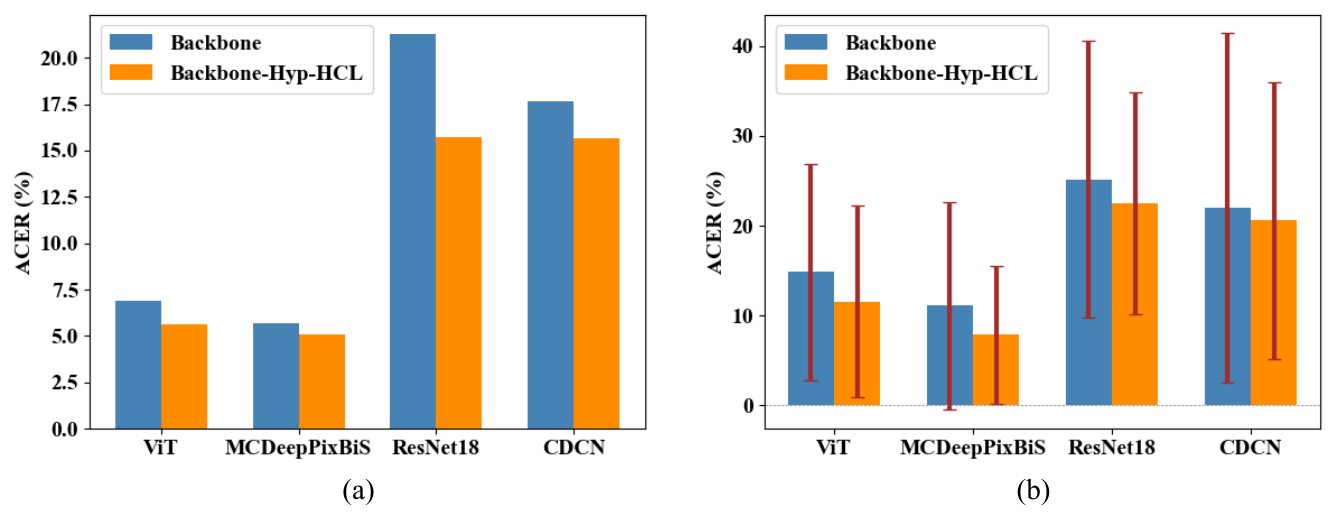}
\caption{Generalization of the framework to other backbones on (a) the WMCA seen protocol; and (b) WMCA unseen protocols.}
\label{fig:abl_model}
\end{figure}

\textbf{Generalization of the unimodal framework to other models on the WMCA dataset.} In order to reflect the versatility of the framework proposed, we use ResNet18 \cite{he2016deep} and CDCN \cite{yu2020searching} as the backbone, and the experimental results on the WMCA seen protocol and unseen protocols are shown in Figure \ref{fig:abl_model}. In this study, only the last fully connected layer of ResNet18 is learnable during training. In the CDCN model, the classification in hyperbolic space is completed based on the low-mid-high levels features extracted by CDCN. It can be found after assembling hyperbolic layers, the performances of ResNet18 and CDCN on both seen and unseen protocols can be improved significantly. This proves that the designed hyperbolic FAS framework has certain versatility.

\textbf{Computational Complexity.} \label{sec:complexity}
The computational and parameter complexity of baselines and proposed hyperbolic unimodal models are shown in \autoref{tab:complexity}. It can be seen that, compared with the MCDeepPixBis model, the FLOPs and the number of parameters on the MCDeepPixBis-Hyp-HCL model only increase by about 0.02M. The FLOPs and the number of parameters on the ViT-Hyp-HCL model also only increase by about 0.09M and 0.1M respectively. Therefore, one of the advantages of the proposed hyperbolic framework is that it only slightly increases the computational complexity while boosting the performance.
\begin{table}[t]
  \centering
  \caption{Computational and parameter complexity}
  \setlength{\tabcolsep}{0.1pt}{
    \begin{tabular}{ccc}
    \textbf{Model} & \textbf{FLOPS} & \textbf{Parameters}\\
    \hline
    ViT   & 11285.49M & 57.30M \\
    \textbf{ViT-Hyp-HCL (ours)} & 11285.58M & 57.40M \\
    \hline
    MCDeepPixBis & 4651.92M & 3.20M \\
    \textbf{MCDeepPixBis-Hyp-HCL(ours)} & 4651.94M & 3.22M \\
    \hline
    \end{tabular}}
  \label{tab:complexity}%
\end{table}%

\begin{table*}[htbp]
\belowrulesep=0ex
\aboverulesep=0ex
  \centering
  \caption{Unimodal experimental results on the MICO benchmarks. The proposed ViT-Hyp-HCL model is compared with state-of-the-art methods in terms of AUC(\%) and HTER(\%). The best results are bolded.}
 \resizebox{0.82\linewidth}{!}{ \begin{tabular}{|c|c|c|c|c|}
    \hline
    \multicolumn{1}{|c|}{\multirow{2}[4]{*}{\textbf{Method}}} & \multicolumn{2}{c|}{\textbf{M\&I to C}} & \multicolumn{2}{c|}{\textbf{M\&I to O}}\\
\cmidrule{2-5} & \multicolumn{1}{c|}{AUC(\%)} & \multicolumn{1}{c|}{HTER(\%)} & \multicolumn{1}{c|}{AUC(\%)} & \multicolumn{1}{c|}{HTER(\%)} \\
    \hline
    ColorTexture~\cite{boulkenafet2016face}   & 46.89 & 55.17& 45.16 & 53.31\\
    \hline
    MS-LBP~\cite{maatta2011face}   & 52.09 & 51.16 & 58.07 & 43.63\\
    \hline
    LBP-TOP~\cite{de2014face}   & 54.88 & 45.27 & 50.21 & 47.26\\
    \hline
    IDA~\cite{wen2015face}   & 58.80 & 45.16 & 42.17 & 54.52\\
    \hline
    MADDG~\cite{shao2019multi}   & 64.33 & 41.02 & 65.10 & 39.35\\
    \hline
    SSDG-M~\cite{jia2020single}   & 71.29 & 31.89 & 66.88 & 36.01\\
    \hline
ViT~\cite{dosovitskiy2020image,liao2023domain}  & 75.86 & 32.34  & 74.42 & \textbf{31.45}\\
    \hline
    \textbf{ViT-Hyp-HCL(ours)} & \textbf{78.81} &\textbf{29.97} & \textbf{75.42}&  32.09 \\ 
    \hline
    \end{tabular}}%
  \label{tab:MM_MICO}%
\end{table*}%

\textbf{Unimodal performance on a limited number of attack types.} We implement the proposed methods on the other four benchmark datasets: MSU-MFSD~\cite{wen2015face}(M),
IDIAP REPLAY-ATTACK~\cite{chingovska2012effectiveness}(I), CASIA-FASD~\cite{zhang2012face}(C), and OULU-NPU~\cite{boulkenafet2017oulu}(O), which only contain print attacks and replay attacks. The experimental results are shown in \autoref{tab:MM_MICO}. It can be seen that the proposed ViT-Hyp-HCL model can surpass the performance of some state-of-the-art methods. Compared with ViT baselines, the hyperbolic FAS framework can bring significant improvement in the M\&I to C protocol. However, in the M\&I to O protocol, ViT-Hyp-HCL performs on par with ViT baselines. This shows that the performance of the hyperbolic framework can be affected on datasets with few types of attacks, which is one of the limitations of our proposed frameworks.

\begin{figure}[t]
\centering
\includegraphics[width=1.0\linewidth]{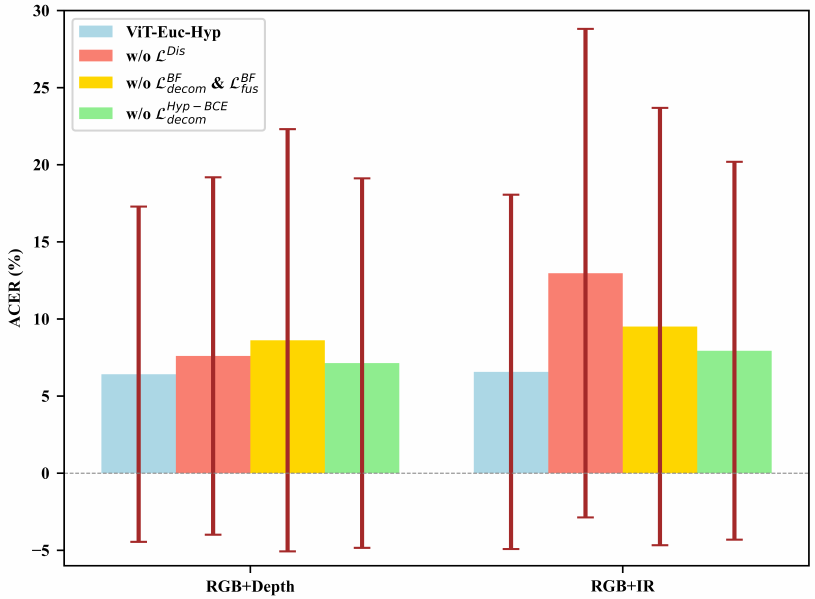}
\caption{Effect of additional losses in the proposed multimodal FAS framework}
\label{fig:MM_loss}
\end{figure}

\textbf{Effect of different kinds of losses in the multimodal FAS framework.} \label{sec:loss_MM} The three kinds of additional losses in the backbone-Euc-Hyp framework are Euclidean distance loss $\mathcal{L}^{Dis}$, hyperbolic contrastive loss $\mathcal{L}^{BF}_{decom}$ \& $\mathcal{L}^{BF}_{fus}$, and hyperbolic cross-entropy loss $\mathcal{L}^{Hyp-BCE}_{decom}$. The effect of these different kinds of losses in the proposed framework is shown in \autoref{fig:MM_loss}. It shows that $\mathcal{L}^{Dis}$, $\mathcal{L}^{BF}_{decom}$, $\mathcal{L}^{BF}_{fus}$ and $\mathcal{L}^{Hyp-BCE}_{decom}$ all have positive impact on the model generalization ability. Among these three losses, $\mathcal{L}^{BF}_{decom}$ and $\mathcal{L}^{BF}_{fus}$ are the most important loss in RGB+Depth modalities and $\mathcal{L}^{Dis}$ is the most effective loss in RGB+IR modalities. Therefore, the three additional losses can supervise the model to improve generalization ability from different aspects.

\begin{figure}[t]
\centering
\includegraphics[width=1.0\linewidth]{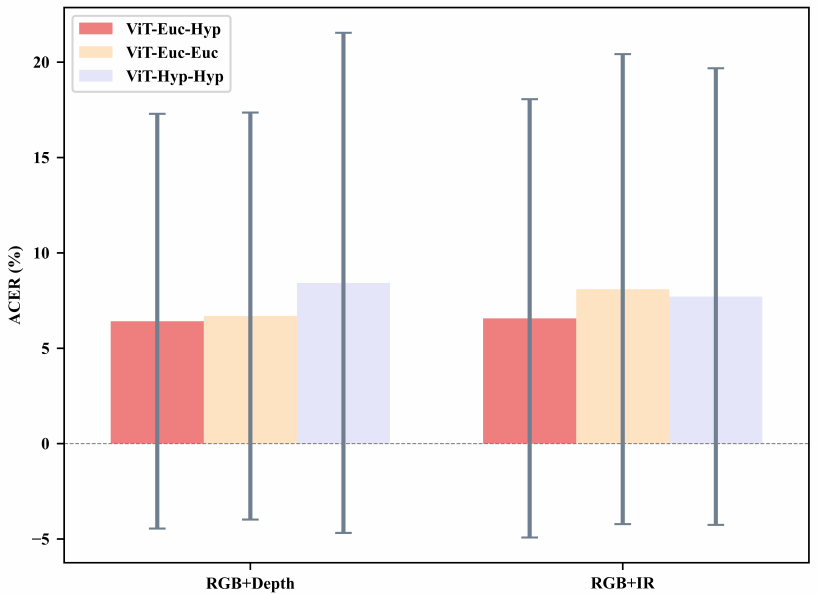}
\caption{Effect of different spaces for multimodal feature decomposition, fusion \& classification.}
\label{fig:MM_space}
\end{figure}%

\textbf{Effect of multimodal feature decomposition in different spaces.} \label{sec:space_MM} In the previous section, we have completed the comparison of multimodal feature fusion \& classification in Euclidean space and hyperbolic space. We want to further explore the effect of multimodal feature decomposition in hyperbolic space. In this case, the unimodal features are mapped into hyperbolic space directly, and the Euclidean distance loss $\mathcal{L}^{Dis}$ is replaced with the hyperbolic distance loss. In addition, it is worth noting that complex hyperbolic concatenation is used to replace simple Euclidean concatenation. The experiment result is shown in \autoref{fig:MM_space}. It can be seen that multimodal feature decomposition in Euclidean space and multimodal feature fusion \& classification in hyperbolic space has the best performance. This is because feature decomposition in hyperbolic space is mathematically complex, especially for hyperbolic concatenation and hyperbolic fully connected layers, which will increase the difficulties of backpropagation. But hyperbolic space is more suitable for multimodal fusion and classification than Euclidean space since embedding in the Poincar\'e ball can provide better generalization ability, as discussed in the previous sections. Therefore, multimodal feature decomposition in Euclidean space followed by fusion \& classification in hyperbolic space is a more efficient combination strategy.

\subsection{Visualization and Analysis} \label{sec:visual}
\begin{figure}[t]
\centering
\includegraphics[width=0.9\linewidth]{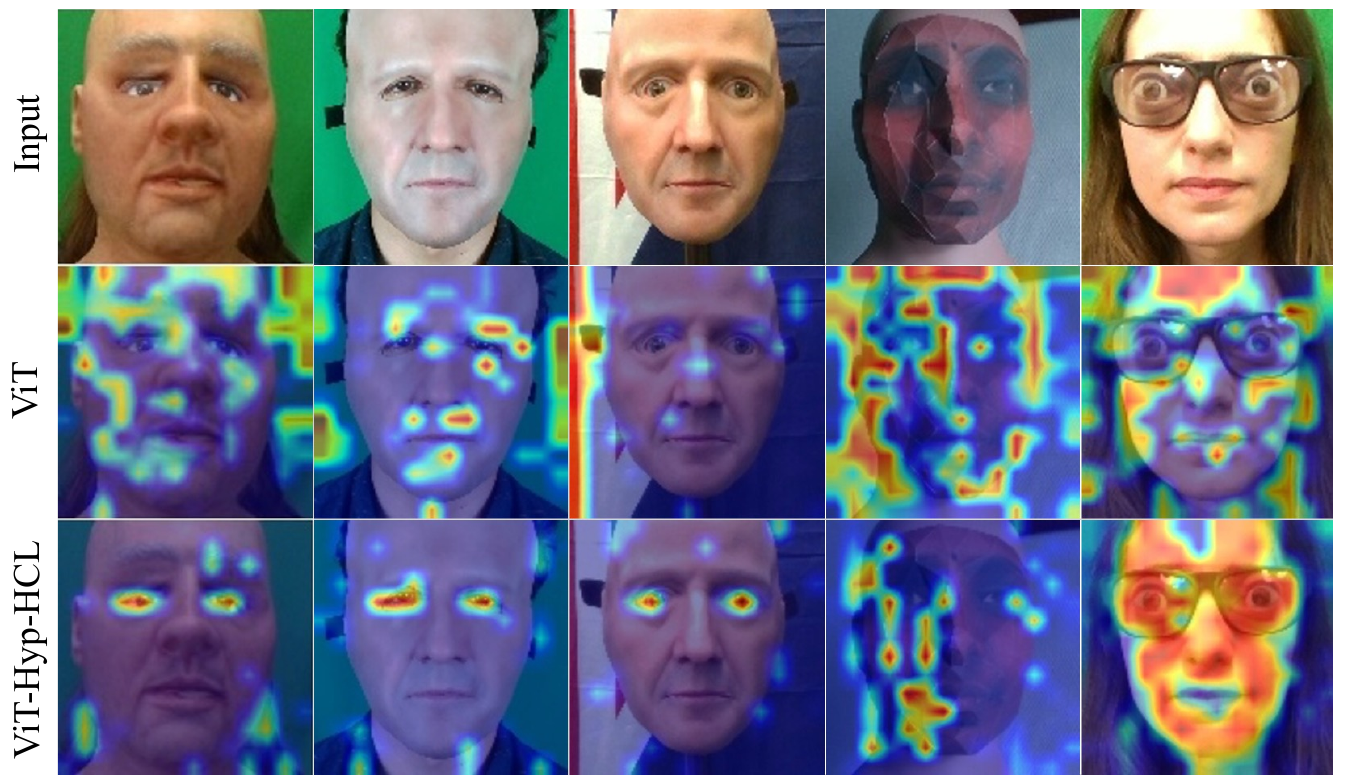}
\caption{Grad-CAM~\cite{selvaraju2017grad} visualization of representative samples on the WMCA dataset.}
\label{fig:grad_cam}
\end{figure}

As shown in Figure \ref{fig:grad_cam}, Grad-CAM~\cite{selvaraju2017grad} is used to reveal the attentional activations of ViT and ViT-Hyp-HCL models for different attack detection. It can be found from the regions of some attack images that the ViT model focuses on are noisy and inefficient. Sometimes it pays too much attention to the background in the attack images for detection. While in the ViT-Hyp-HCL model, the flaws in the attack image can be found more accurately. For example, in the first to third columns of Figure \ref{fig:grad_cam}, the ViT-Hyp-HCL model perceives the unreality in the eyes and the region around the eyes, while the baseline ViT model seems clueless. This elucidates that the model based on hyperbolic space is able to mine explainable spoofing cues to detect attacks.

\section{Conclusions and Future Work}
\label{sec:conc}
In this paper, a novel hyperbolic contrastive learning framework for unimodal face anti-spoofing (FAS) is proposed. And a new feature clipping method is used to alleviate the vanishing gradient problem in hyperbolic neural network training. Then, a novel multimodal FAS framework consisting of Euclidean multimodal feature decomposition and hyperbolic multimodal feature fusion \& classification is designed. Both the proposed hyperbolic unimodal and multimodal frameworks can bring improvement on intra- and cross-dataset testings of WMCA, PADISI-Face, SiW-M, MSU-MFSD, IDIAP REPLAY-ATTACK, CASIA-FASD, and OULU-NPU datasets, which reflect their versatility and generalization ability. However, we note that research on FAS in hyperbolic space is still in its infancy. Only some basic properties and operations of hyperbolic space are used in our work while the full potential of hyperbolic space on FAS is encouraged to exploit in the future.

\section{Data Availability Statement}
Experimental data from seven datasets: WMCA \cite{george2019biometric}, PADISI-Face \cite{rostami2021detection,spinoulas2021multispectral}, SiW-M \cite{DTL}, MSU-MFSD~\cite{wen2015face}, IDIAP REPLAY-ATTACK~\cite{chingovska2012effectiveness}, CASIA-FASD~\cite{zhang2012face}, and OULU-NPU~\cite{boulkenafet2017oulu}). The data that support the findings of this study are available from third-party institutions (including Idiap Research Institute, USC Viterbi School of Engineering Information Sciences Institute, Michigan State University, Institute of Automation, Chinese Academy of Sciences, OULU-NPU Database) but restrictions apply to the availability of these data, which were used under license for the current study. Data are however available from the authors upon reasonable request and with permission of the above-mentioned third-party institutions.

\bibliography{Hyperbolic_FAS_submission}

\end{document}